\documentclass{article}

    \usepackage[preprint]{neurips_2025}

\usepackage[utf8]{inputenc} 
\usepackage[T1]{fontenc}    
\usepackage{hyperref}       
\usepackage{url}            
\usepackage{booktabs}       
\usepackage{amsfonts}       
\usepackage{nicefrac}       
\usepackage{microtype}      
\usepackage{xcolor}         
\usepackage{multirow}

\usepackage{graphicx}
\usepackage{subfigure}
\usepackage{todonotes}
\usepackage{algorithm}
\usepackage{algpseudocode}

\usepackage{amsmath}
\usepackage{amssymb}
\usepackage{mathtools}
\usepackage{amsthm}

\usepackage{xfrac}
\usepackage{subcaption}
\usepackage{wrapfig}
\usepackage{enumitem}
\usepackage{array}

\usepackage{makecell}

\theoremstyle{plain}
\newtheorem{theorem}{Theorem}[section]

\theoremstyle{definition}

\theoremstyle{remark}

\definecolor{color1}{HTML}{1f77b4}
\definecolor{color2}{HTML}{ff7f0e}
\definecolor{color3}{HTML}{2ca02c}
\definecolor{color4}{HTML}{e4bcad}
\definecolor{color5}{HTML}{76c8c8}
\definecolor{color6}{HTML}{c80064}
\definecolor{color7}{HTML}{3c4e4b}
\definecolor{color8}{HTML}{7f7f7f}
\definecolor{color9}{HTML}{bcbd22}
\definecolor{color10}{HTML}{17becf}
\definecolor{color11}{HTML}{aec7e8}
\definecolor{color12}{HTML}{ffbb78}
\definecolor{color13}{HTML}{98df8a}
\definecolor{color14}{HTML}{ff9896}
\definecolor{color15}{HTML}{c5b0d5}
\definecolor{color16}{HTML}{c49c94}
\definecolor{color17}{HTML}{f7b6d2}
\definecolor{color18}{HTML}{c7c7c7}
\definecolor{color19}{HTML}{dbdb8d}
\definecolor{color20}{HTML}{9edae5}
\definecolor{color21}{HTML}{ad494a}

\usepackage{pgfplots}
\usepackage{adjustbox}
\usepgfplotslibrary{fillbetween}  
\usetikzlibrary{arrows.meta}
\tikzset{
    myarrow/.style={-{Triangle[length=1.5mm,width=1.5mm]}}
}

\title{The Quest for Winning Tickets in Low-Rank Adapters}

\author{%
  \begin{tabular}{c}
    \textbf{Hamed Damirchi$^{1}$ \qquad Cristian Rodriguez-Opazo$^{1}$ \qquad Ehsan Abbasnejad$^{2}$} \\
    \textbf{Zhen Zhang$^{1}$ \qquad Javen Shi$^{1}$} \\ [1ex]
    \textnormal{$^{1}$Australian Institute for Machine Learning, Adelaide University \qquad $^{2}$Monash University} \\
    \textnormal{\texttt{\{firstname.lastname\}@adelaide.edu.au}}
  \end{tabular}
}

\begin{document}

\maketitle

\begin{abstract}
The Lottery Ticket Hypothesis (LTH) suggests that over-parameterized neural networks contain sparse subnetworks ("winning tickets") capable of matching full model performance when trained from scratch. With the growing reliance on fine-tuning large pretrained models, we investigate whether LTH extends to parameter-efficient fine-tuning (PEFT), specifically focusing on Low-Rank Adaptation (LoRA) methods. 
Our key finding is that LTH holds within LoRAs, revealing sparse subnetworks that can match the performance of dense adapters. 
In particular, we find that the effectiveness of sparse subnetworks depends more on how much sparsity is applied in each layer than on the exact weights included in the subnetwork.
Building on this insight, we propose Partial-LoRA, a method that systematically identifies said subnetworks and trains sparse low-rank adapters aligned with task-relevant subspaces of the pre-trained model. Experiments across 8 vision and 12 language tasks in both single-task and multi-task settings show that Partial-LoRA reduces the number of trainable parameters by up to 87\%, while maintaining or improving accuracy. Our results not only deepen our theoretical understanding of transfer learning and the interplay between pretraining and fine-tuning but also open new avenues for developing more efficient adaptation strategies.
\end{abstract}

\section{Introduction}
\label{sec:intro}

Fine-tuning large pre-trained models for downstream tasks has become a cornerstone of modern machine learning,
unlocking their potential across diverse applications in vision, language and beyond.
However, traditional fine-tuning methods require updating all model parameters, which is computationally expensive, energy-intensive, and impractical for resource-limited environments.
Moreover, alternative approaches to fine-tuning still rely on computationally expensive techniques, 
such as specialized layers \citep{xu2023side}, adjusting normalization \citep{giannou2023expressive}, retraining or introducing parallel layers \citep{chen2022adaptformer}.

To address this, Parameter-Efficient Fine-Tuning (PEFT) methods \citep{xin2024parameterefficient} have been developed. 
These approaches aim to adapt large models to downstream tasks without the computational overhead of training all model parameters.
PEFT methods achieve this by introducing a limited number of new trainable parameters while keeping the base model frozen.
Among these, 
Low-Rank Adaptation (LoRA) \citep{hu2022lora} has emerged as a leading approach. 
LoRA obtains a balance between efficiency and performance
for model fine-tuning by introducing low-rank matrices,
that adjust base model weights during training.
This makes LoRA particularly suitable for
large model adaptation
while maintaining scalability and affordability.

Despite the success of LoRA, a fundamental inefficiency remains: even with low-rank matrices, LoRA modifies every element of the original weight matrix. This overlooks the \textbf{low intrinsic dimensionality} of large pretrained models \citep{aghajanyan2020intrinsic, hu2022lora}. At the same time, the \textbf{lottery ticket hypothesis (LTH)} \citep{frankle2019lth} suggests that only a small, carefully chosen subnetwork, known as a ``\emph{winning ticket}'', may be sufficient to match the performance of the full model when trained in isolation. Yet existing PEFT methods, including LoRA, do not explicitly leverage this sparsity, leaving untapped potential for efficiency and sparsity in fine-tuning.

\begin{figure*}[t]
    \centering
    \includegraphics[width=0.97\textwidth]{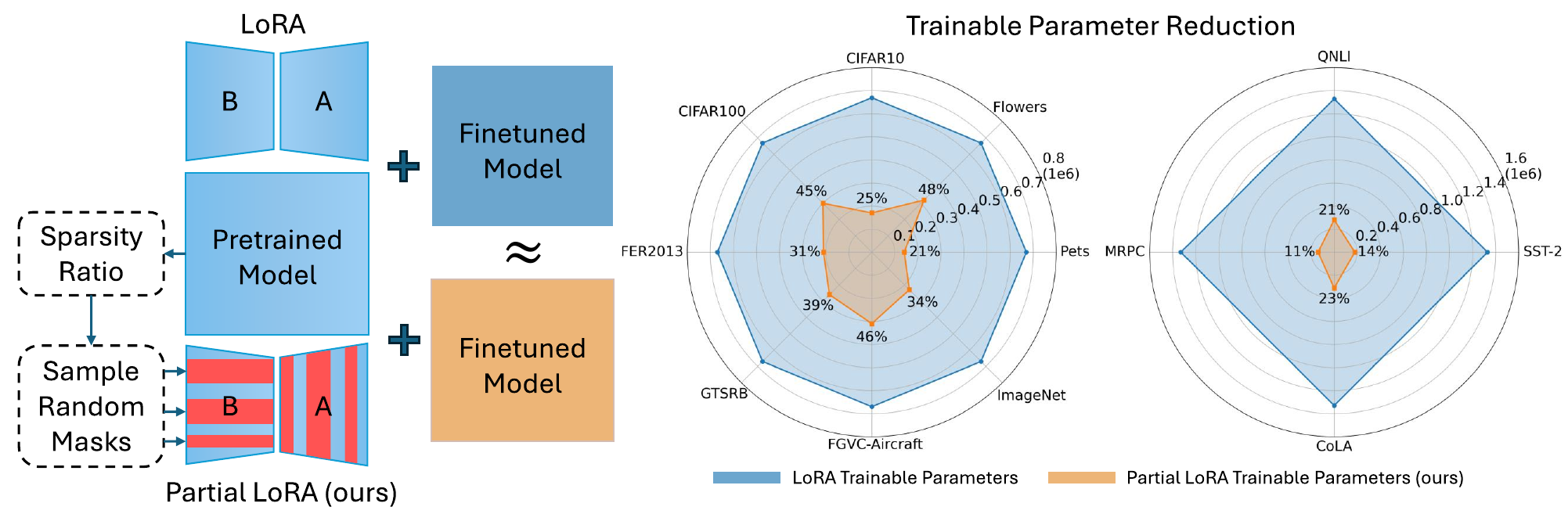}
    \caption{The process of sparsifying LoRAs (left) involves 
    extracting sparsity ratios from the pre-trained model to generate random masks for low-rank adapter components, yielding outputs comparable to fully-parameterized LoRA while substantially reducing trainable parameters, as demonstrated for vision and language tasks (right).}

    \label{fig:teaser}
    \vspace{-8mm}
\end{figure*}

To address these limitations, 
we propose extending the theoretical framework of the lottery ticket hypothesis and random masking \citep{gadhikar_why_2023} to LoRA. 
Specifically, 
we demonstrate that randomly masking the weights of a low-rank adapter in each layer can yield tight bounds on adapter outputs while preserving downstream task performance. 
The sparsity of these masks is dynamically determined 
based on the model's capacity requirements for the target task. 
Our findings reveal that LoRA residuals can be substantially sparsified, 
similar to the effects observed in random pruning of full models \citep{chijiwa_pruning_2022}. 
Crucially, this approach confirms that fine-tuning can rely on "winning tickets" within LoRAs, 
where any randomly masked low-rank adapter with the same sparsity can achieve performance comparable to the fully parameterized LoRA.
Building on these theoretical insights, 
we introduce Partial-LoRA, 
a method to identify and apply task-specific sparsity ratio for fine-tuning models. 
Partial-LoRA derives per-layer sparsity ratios and limits fine-tuning to a subset of weights by relating top subspaces of pre-trained weights to specific matrix elements (Figure \ref{fig:teaser}).
Unlike traditional pruning methods, 
which require gradient-based importance measures \citep{lee_snip_2019, imp2019molchanov}, our approach leverages the structure of pre-trained models to extract sparsity ratios without additional overhead. 
By randomly masking weights according to these ratios, 
Partial-LoRA achieves a significant reduction in trainable parameters 
while preserving performance across a wide range of tasks.

We validate the effectiveness of Partial-LoRA through extensive experiments on visual and language tasks in both single-task and multi-task settings. 
Our results demonstrate that random masking, according to the derived sparsity ratios, leads to a significant reduction in the number of parameters, with up to \textbf{87\% fewer parameters} needed for residuals compared to LoRA, 
while maintaining or even improving performance in some cases.

Furthermore, we compare the random masking approach with deterministic pruning techniques, showing that \textbf{Partial-LoRA achieves superior performance} on average. 
This shows that the precise selection of sub-networks may be less important than maintaining the overall sparsity within a task-specific limit.
Our main contributions are as follows.
\begin{itemize}[topsep=-2pt,itemsep=-1pt,labelindent=2pt,labelsep=*,leftmargin=12pt]

\item We extend the lottery ticket hypothesis (LTH) to LoRAs, showing that randomly masked LoRAs can achieve performance comparable to dense LoRAs, even with substantial sparsification.

\item We show that precise sub-network selection is not essential. With sufficient elements, LoRAs can effectively adjust key subspaces of the pre-trained weight matrix, ensuring efficient fine-tuning. 

\item Our experiments show that random masking of LoRAs can reduce parameters by up to 87\% with minimal performance loss, even outperforming full LoRAs in some cases. For ViT-B-16, trainable parameters are reduced from 110M to 120k, enabling fine-tuning and inference without specialized hardware with robust performance.

\end{itemize}
\section{Related Work}
\label{sec:relatedwork}

\textbf{Parameter Efficient Fine-Tuning (PEFT)} methods adapt large models with significantly fewer trainable parameters than full fine-tuning. 
Low-Rank Adaptation (LoRA) \citep{hu2022lora}, uses trainable low-rank matrices added at each layer to achieve parameter efficiency.
Extensions such as AdaLoRA \citep{zhang2023adaptive} dynamically adjust the rank of residual matrices during training.
Other works, including DoRA \citep{liu2024dora} and LoRA+ \citep{hayou2024lora}, address limitations in LoRA's training dynamics, aiming to bridge the gap between low-rank and full-rank fine-tuning.
Our method builds on LoRA and AdaLoRA while complementing other advancements, offering efficient fine-tuning with task-adaptive sparsity.

\textbf{Pruning and Layer-specific Adjustments.}
Recent efforts like LoRAPrune \citep{zhang2023loraprune} and LoRA-Shear \citep{Chen2023LoRAShearEL} introduce pruning techniques targeting LoRA modules.
LoRAPrune estimates importance using LoRA weights and gradients, while LoRA-Shear employs structured pruning guided by dependency graphs.
However, these approaches focus on pruning the original model, not masking LoRAs. 
Alternatively, PRILoRA \citep{benedek-wolf-2024-prilora} tailors layer-specific ranks but lacks parameter reduction guarantees.
Our approach complements such methods by enabling random masking within layer-specific LoRA ranks, combining efficiency with adaptability.

\textbf{Parameter Reduction.}
IncreLoRA \citep{Zhang2023IncreLoRAIP} decomposes adapters into rank-1 modules, dynamically adding ranks based on task needs while limiting overparameterization.
VeRA \citep{kopiczko2024vera} shares low-rank matrices across layers, reducing trainable parameters but potentially increasing overall memory requirements.
Our approach differs by employing random masking informed by pretrained model structure, 
reducing parameters without increasing computational overhead.
A closely related work is \citet{xu2024randommaskingfindswinning}, which uses static random masking of LoRA residuals.
Unlike this approach, our method determines sparsity adaptively for each layer, guided by the model, allowing for principled task-specific capacity allocation.

\textbf{Theorical Insights.}
\citet{gadhikar_why_2023}, building up on the previous work from \citet{burkholz_existence_2022} and \citet{burkholz2022most}, proved the existence of strong lottery tickets in randomly masked variants of large overparameterized models.
We extend these results to low-rank adaptation. 
Notably, \citet{gadhikar_why_2023} demonstrated that a target model could be approximated using only $L+1$ layers from the original model. Our work extends this proof to show that the same principles hold for masked LoRAs, providing a theoretical foundation for maintaining approximation quality while reducing parameter count.

\section{Background}
\label{sec:background}
Given a model $M$ parameterized by ${W, b}$ with depth $L$, LoRA \citep{hu2022lora} fine-tunes the weight matrix $\mathbf{W}^l \in \mathbb{R}^{m \times n}$ at layer $l$ by introducing a low-rank residual trainable matrix $\Delta \mathbf{W}_l \in \mathbb{R}^{m \times n}$. This residual is added to the weight matrix $\mathbf{W}_l$ of the frozen pre-trained model $M$. While no bias terms are added to the layer, the only trainable parameters are the residuals added at each layer. This results in each layer of the new model being formulated as:
\begin{equation}
    \mathbf{h}_l = \sigma((\mathbf{W}_l+\Delta \mathbf{W}_l)\mathbf{x}+\mathbf{b}_l)\,,
\end{equation}
where $\mathbf{h}_l$ is the output of the layer and $\mathbf{b}_l$ represents the bias term at layer $l$ from the original pre-trained model. $\sigma$ and $\mathbf{x}$ represent the layer nonlinearity and input, respectively. The matrix $\Delta \mathbf{W}_l$ itself is defined as:
\begin{equation}
    \Delta \mathbf{W}_l = \mathbf{B}_l\mathbf{A}_l,\enspace \mathbf{B}_l \in \mathbb{R}^{m \times d},\, \mathbf{A}_l \in \mathbb{R}^{d \times n}\,,
\end{equation}
where $\mathbf{B}_l$ and $\mathbf{A}_l$ are low-rank trainable matrices. Using this formulation, the number of trainable parameters is reduced from $m\times n$ to $m\times d+n\times d$ where $d\ll\min(m, n)$.
\section{Methodology}
\label{sec:methodology}

Here, we outline our work divided into two key sections. Section \ref{sec:lth} extends Strong Lottery Ticket theory to Low-Rank Adapters (LoRAs), showing that randomly masked LoRAs can approximate a target adapter if the unmasked adapter is logarithmically wider. 
In Section \ref{sec:subnetwork} we detail the extraction of LoRA subnetworks (LoRA masks) using limited labeled data by pruning adapter weights based on their importance in the pre-trained model.
\subsection{Existence of Lottery Tickets in LoRA Residuals}
\label{sec:lth}

In this section, we extend the existing theoretical work on Strong Lottery Tickets (SLT) \citep{gadhikar_why_2023} to show that a randomly pruned LoRA can approximate a target LoRA (winning ticket), provided the unpruned LoRA is wider by a logarithmic margin. In other words, the pruned LoRA represents a subnetwork of the fully parameterized one, capable of approximating a target network.
\begin{theorem}
Define a network \( f_{T} \) of depth \( L \), parameterized by pretrained weights and biases \( W_l \) and \( b_l \), with low-rank adapters at each layer parameterized by residuals \( \Delta W^l_T \). Additionally, define a pruned network \( f_{LoRA} \) of depth \( L+1 \), parameterized by \( \Delta W^l \cdot U \), where \( U\sim B(p^l) \) is a mask sampled from a Bernoulli distribution, and the same \( W_l \), \( b_l \) as the target model. This pruned model consists of sparsity factors \( p_l \) at layer \( l \), and the residuals \( \Delta W^l_{ij} \sim U([-1, 1]) \). Both networks have \( n_{T, l} \) and \( n_{LoRA, l} \) neurons at layer \( l \). 
Then, given variables \(\epsilon, \delta \in (0, 1)\), with failure probability \( 1 - \delta \), there exists a mask \( U \) such that, for all \( x \in \mathcal{D} \) within the compact space \( \mathcal{D} \), it holds that $\max_{x\in \mathcal{D}}||f_{T}(x; \Delta W_T)-f_{LoRA}(x; \Delta W \cdot U)||\leq \epsilon$ if:
\label{th:theorem}
\end{theorem}
\begin{equation}
    n_{LoRA,l} \geq C\frac{n_{T,l}}{\log(\sfrac{1}{1-p_{l+1}})}\log(\frac{1}{\min(\epsilon_l, \sfrac{\delta}{\rho})}).
\end{equation}
Where $C$ is a distribution dependent constant and $\rho$ and $\epsilon_l$ are defined following \citet{gadhikar_why_2023} and \citet{burkholz2022most}. The proof, provided in Appendix \ref{app:proof},
hinges on the idea that after multiplying the neurons in the target adapter by a certain factor, we can ensure that at least one edge remains unpruned after pruning the original LoRA, preserving the adapter's functionality. We show that changes to each layer induced by LoRAs do not alter the proof in \citet{gadhikar_why_2023} and \citet{burkholz2022most}.  

A concern with this masking approach is flow preservation \citep{burkholz2022most}. In some randomly pruned models, two consecutive layers can end up with mismatched pruned weights, causing any input to those layers to result in a zero output due to a discontinuity. This happens when non-zero outputs from the first layer are multiplied by pruned elements in the second layer. To address this, a common modification is to adjust the non-zero indices to maintain input flow. However, in this work, this step is unnecessary because the pretrained model's frozen layers remain unchanged, ensuring that flow is preserved even if the low-rank residual weights are set to zero.

Theorem \ref{th:theorem} shows that given a full-parameter adapter, there exists a randomly masked adapter dependent on layer width capable of delivering performance that matches the full-parameter adapter while using fewer parameters. In practice, we show that while the exact ratio of this reduction in the number of parameters is unknown beforehand (since the target LoRA is unknown), importance measuring criterion from the pruning literature (Section \ref{sec:importance}) may be used as a proxy to obtain this ratio. We name this ratio the capacity at each layer required to learn the task at hand. 

\subsection{Finding Subnetworks in Pretrained Models}
\label{sec:subnetwork}
To extract sparsity ratios from pretrained models, we propose a two-step approach. First, we present an algorithm for deriving sparsity ratios based on the model's performance on a small subset of labeled instances. Then, we discuss various importance measures that can be used to rank the significance of elements in the weight matrix. This section details both the algorithmic approach (Section \ref{subsec:algorithm}) and the importance measures (Section \ref{sec:importance}) used in our method.

\subsubsection{Deriving Sparsity Ratios}
\label{subsec:algorithm}
We start by randomly sampling a small subset of labeled instances from the dataset in a few-shot scenario. This subset, $\mathcal{D}_t$, with $m$ samples from the dataset $\mathcal{D}$, is used to measure the pre-trained model's accuracy as a baseline. Importance measures are then employed to rank the significance of each element in the weight matrix. The process for importance measures is detailed in Section \ref{sec:importance}. Starting with the least important elements, the weights are progressively masked until the model's accuracy drops below 90$\%$ of the baseline, using the same few-shot dataset $\mathcal{D}_t$. This specific accuracy margin is adopted from the literature on intrinsic dimensionality \citep{aghajanyan2020intrinsic, Li2018MeasuringTI} where the 90\% margin is deemed a satisfactory solution for fine-tuning a pre-trained model using only a small subset of parameters (the intrinsic dimension) while maintaining most of its effectiveness. This iterative masking is applied to the pre-trained model, yielding a sparsity ratio for each layer, as outlined in Algorithm \ref{alg:subnetwork}. Random masks are then generated using these ratios by sampling from a Bernoulli distribution. During low-rank fine-tuning, only the unmasked elements of weight matrices are modified at each layer.

\begin{algorithm}[t]
\caption{Sparsity ratio derivation for a single layer}
\label{alg:selective_finetuning_single_layer}
\begin{algorithmic}[1]
\Statex Pretrained model $M$, weight matrix $\mathbf{W}^l$ for layer $l$ of $M$, input-output data pairs $(\mathbf{x}, \mathbf{y})$ forming dataset $D_t$, where $D_t \subset D$ consists of $m$ shots randomly sampled per class from $D$
\State $\mu\gets \frac{1}{N} \sum_{(x,y)\in D_t} \mathbb{I}[M(\mathbf{x}, \mathbf{W}) = y]$ \Comment{Determine few-shot accuracy on $D_t$}
\State $\mathbf{\hat{y}} \gets M(\mathbf{x})$ \Comment{Compute model output for $\mathbf{x}\in D_t$}
\State $I^l \gets \text{importance}(\mathbf{\hat{y}}, \mathbf{y}, \mathbf{W}^l)$ \Comment{Compute importance scores for every element of $\mathbf{W}^l$}
\State $\mathbf{U} \gets \mathbf{1}$ \Comment{Initialize mask matrix with ones}
\While{$0.9\times\mu \leq \frac{1}{N} \sum_{(\mathbf{x},y)\in D_t} \mathbb{I}[M(\mathbf{x}, \mathbf{W}^l \odot \mathbf{U}) = y]$}
\State $I^l \gets \text{importance}(\mathbf{W}^l, D_t)$
\State $\mathbf{U}_{ij} \gets 0$ for all $(i, j) \notin I^l$ \Comment{Update mask matrix to zero out less important weights}
\EndWhile
\State \Return $|I^l|$
\end{algorithmic}
\label{alg:subnetwork}
\end{algorithm}

\subsubsection{Importance Measures}
\label{sec:importance}
Commonly used importance measures \citep{lee_snip_2019, imp2019molchanov} focus on an element-level computation deriving a scalar value for each element in a weight matrix. However, based on the formulation of low-rank adaptation detailed in Section \ref{sec:background}, it is enough to infer the important rows and columns of the weight matrix and there is no necessity to focus on specific elements when dealing with LoRAs. This is because for every element of the residual we have $\Delta w_{ij}=\sum_{k}^{d} \mathbf{b}_{i,k}\mathbf{a}_{k, j}$. Therefore, rather than masking the residual weights $\Delta \mathbf{W}=\mathbf{BA}$ we can opt to instead mask $\mathbf{B}$ and $\mathbf{A}$ separately which removes any requirement for computing a large matrix that will be masked before the forward pass of the inputs. This way, the unimportant rows of $\mathbf{W}$ will lead to masking rows of $\mathbf{B}$ while unimportant columns will lead to masking $\mathbf{A}$. 

To this end, we calculate \textbf{leverage scores} to identify the rows and columns most critical to preserving the top-$k$ subspace of the weight matrix. We compute the truncated singular value decomposition $\mathbf{W} \approx \mathbf{P}_k\mathbf{\Lambda}_k \mathbf{Q}_k$, where $\mathbf{P}_k$ and $\mathbf{Q}_k$ contain the top-$k$ left and right singular vectors, respectively. We then define the importance of each row and column as the squared $L_2$ norm of the corresponding row vectors in the singular matrices. This metric specifically measures the alignment of each input and output dimension with the principal components of the layer. Thus, we derive the row leverage scores as:
\begin{equation}
    I^l_{row, i} = \sum_{j=1}^k (\mathbf{P}_{ij})^2 = ||(\mathbf{P}_k)_{i,:}||_2^2,
\end{equation}
where $I^l_{row, i}$ represents the leverage score for the $i$-th row. We apply the same logic to $\mathbf{Q}_k$ to compute the column leverage scores. We use these leverage scores as the importance criterion in Algorithm \ref{alg:subnetwork} to select the indices that maximally preserve the pretrained features. Alternatively, any importance measure in the pruning literature such as SNIP \citep{lee_snip_2019} or IMP \citep{imp2019molchanov} may be used. SNIP, as a single-shot pruning approach, uses the absolute value of the gradient of the output with respect to the weight matrix at every layer as a proxy for the importance which makes the method dependent on the pairs of samples passed through the network. IMP, on the other hand, scales the gradients by the weights to also consider the magnitude of weights alongside their gradient. Both of these approaches compute the importance in an element-wise manner. Therefore, to derive the indices required for adaptation, we can take the row-wise or column-wise sum over the importance values. Regardless of the importance criterion, to derive the mask $\mathbf{U}$ in Algorithm \ref{alg:subnetwork}, we sample row and column masks $\mathbf{u}_{row}$ and $\mathbf{u}_{col}$ according to $|I^l|$ obtained from Algorithm \ref{alg:subnetwork} where:
\begin{equation}
    \Delta \mathbf{W} = (\mathbf{u}_{row} \odot \mathbf{B})(\mathbf{A} \odot \mathbf{u}_{col})\,.
    \label{eq:maskingform}
\end{equation}
Here, $\mathbf{u}_{row} \sim \text{Bernoulli}(p_{row})$ and $\mathbf{u}_{col} \sim \text{Bernoulli}(p_{col})$, where $p_{row}$ and $p_{col}$ are determined based on the ratio of $|I^l|$ over the row or column size of $\mathbf{W}$. We further refer to Appendix \ref{app:mask_decomp} for an analysis of how the masking formulation in (\ref{eq:maskingform}) allows for the results of Theorem \ref{th:theorem} to hold.

Therefore, our approach, which we name \textbf{Partial-LoRA}, leverages importance measuring criteria to determine a minimal capacity required at each layer for fine-tuning pre-trained models on downstream tasks with LoRAs. By using these importance scores as a proxy for the sparsity ratios established in Theorem \ref{th:theorem}, we can generate random masks that maintain bounded output differences between masked and unmasked LoRAs.

\section{Experiments}
\label{sec:experiments}
\begin{figure*}[tbp!]
    \centering
    \includegraphics[width=0.9\linewidth]{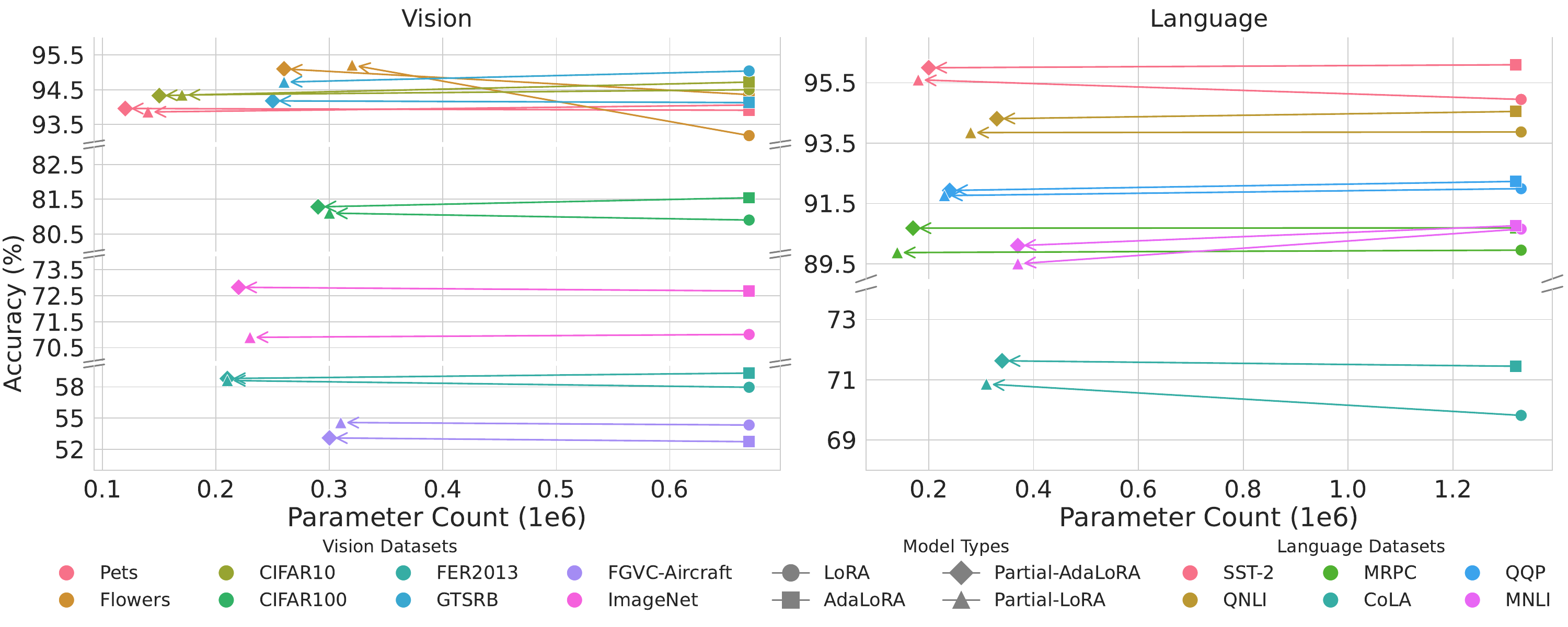}
    \vspace{-1.5mm}
    \caption{Change in accuracy is visualized against parameter count (in millions) after the application of our sparsification method for both vision and language models. Both Partial-AdaLoRA and Partial-LoRA performance remains consistent after significant sparsification.}
    \label{fig:sparsifying_vis}
    \vspace{-4mm}
\end{figure*}
\textbf{Datasets and Models.} We use OpenAI CLIP for our image classification experiments, adding a linear classifier on top of the image features and initializing it with language prompts. We use OxfordIIITPet (Pets) \citep{parkhi12a}, CIFAR-10 \citep{Krizhevsky2009LearningML}, CIFAR-100 \citep{Krizhevsky2009LearningML}, Flowers \citep{flowers}, FER2013 \citep{fer2013}, GTSRB \citep{gtsrb}, FGVC-Aircraft \citep{fgvcaircraft} and ImageNet-1k \citep{ILSVRC15} datasets, with respective class counts of 37, 10, 100, 101, 7, 43, 102, and 1000 representing tasks with varying levels of complexity. Each dataset uses 16 shots, except Flowers, using 6 shots due to its smaller size. We employ a similar validation set for early stopping. The same few-shot sets are used for both subnetwork extraction and training.

We use LLAMA2-7b \citep{llama}, and Deberta-V3-Base \citep{he2021debertav3} for our language model tests. For Deberta-V3-Base, we perform experiments similar to vision on the GLUE benchmark \citep{glue}. To determine sparsity ratios, 2\% of the smaller and 5\% of the larger datasets are used. While a few-shot dataset is used for deriving the ratios, the whole dataset is used for training to match the SotA workflow. For LLAMA2, we use WSC \citep{levesque2012winograd}, SST-2 \citep{glue}, and COPA \citep{Gordon2011ChoiceOP} with a few-shot train set for single-task experiments. For multi-task learning, we use BoolQ \citep{clark2019boolq}, COPA, RTE \citep{glue}, and WiC \citep{superglue}.
Models are trained with the AdamW optimizer and a cosine annealing learning rate scheduler. We train with varying seeds and initial learning rate range of \texttt{\{0.0001-0.005\}}, resulting in 20 models per method, and report the mean performance of the top 5 based on validation scores. For GLUE, we report results on the development set following \citet{zhang2023adaptive}.
\begin{figure*}[h]
    \centering
        \includegraphics[width=0.99\linewidth]{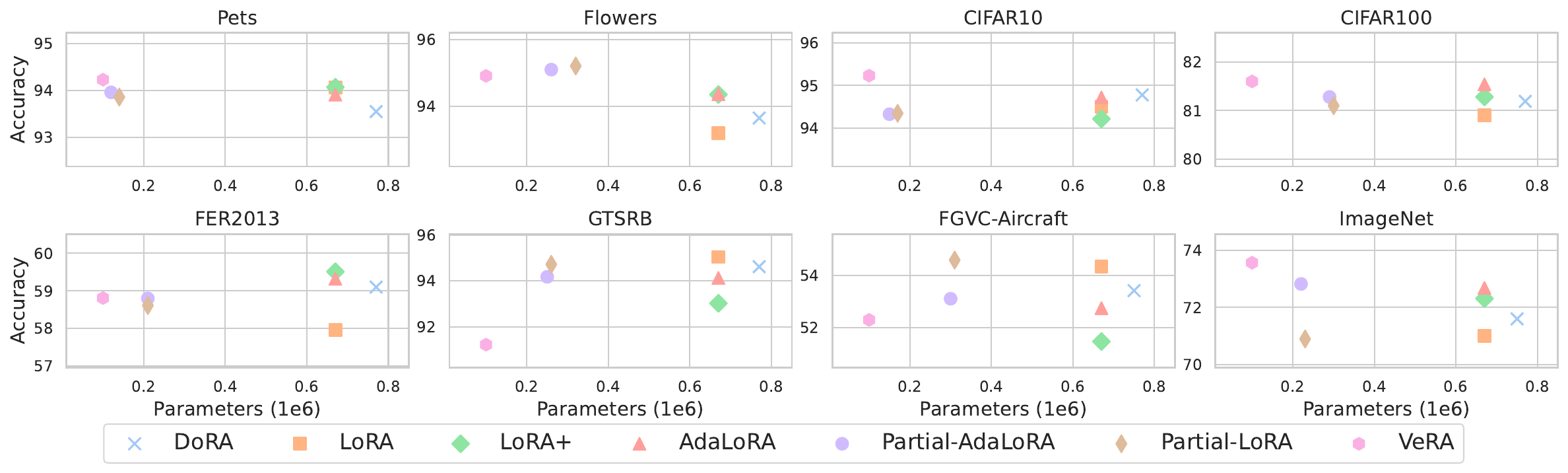}
        \caption{Despite the fewer parameters of our Partial method, our results are competitive with SotA adaptation methods on vision datasets with improvements in datasets prone to overfitting (Flowers).}
        \label{fig:sota_vis}
        \vspace{-3mm}
\end{figure*}
\\
\textbf{Setting and Baselines.} We apply our method to both LoRA~\citep{hu2022lora} and AdaLoRA~\citep{zhang2023adaptive} resulting in Partial-LoRA and Partial-AdaLoRA, demonstrating the potential for sparsification in any subsequent methods based on LoRA. In addition to AdaLoRA, we also compare against LoRA+, DoRA, and VeRA as SotA PEFT methods and use Pyramidal \citep{liu_unreasonable_2022} and Balanced \citep{frankle2019lth} random pruning methods as baselines. The Pyramidal approach sets layer sparsity as $p_l=p^l$ where $p$ is first-layer sparsity and $l$ is layer depth, while Balanced uses fixed sparsity across layers. Though these methods lack principled sparsity derivation and ignore each layer's requirements, comparing their accuracy across sparsity levels gives valuable insights.
\subsection{Sparsifying LoRAs}
\label{sec:performance}
Here we analyze the impact of sparsifying LoRA and AdaLoRA, using Algorithm \ref{alg:subnetwork}, using our SVD-based approach. As shown in Figure \ref{fig:sparsifying_vis}, our method significantly reduces the number of trainable parameters across both vision and language tasks. For vision, we observe an average reduction of over 60\%, with a maximum of 80\% for the Pets dataset. Language tasks show even greater reductions, averaging 82\% with a maximum of 87\% for MRPC. Importantly, these reductions are achieved while maintaining accuracy compared to the unmasked LoRA and AdaLoRA models across both modalities. An exception is the Flowers dataset where we see a significant increase in accuracy which we discuss later in Section \ref{sec:sota_com}. 
Although the results in Figure \ref{fig:sparsifying_vis} are obtained through a few-shot dataset, we also report full dataset results in Appendix \ref{sec:wholedataset}, demonstrating consistent performance patterns.
\begin{figure*}[t!]
    \centering
    \includegraphics[width=0.99\linewidth]{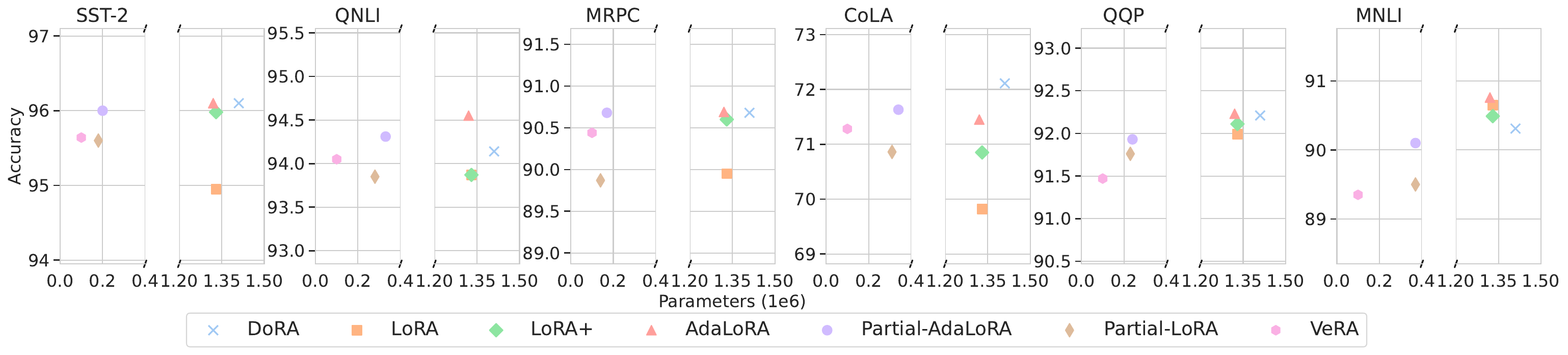}
    \vspace{-1.5mm}
    \caption{Partial-LoRA and Partial-AdaLoRA stay competitive with SotA methods on language datasets despite the low parameter count.}
    \label{fig:sota_lang}
    \vspace{-5mm}
\end{figure*}
\subsection{Comparison to the SoTA}
\label{sec:sota_com}
We compare the results of sparsified LoRA and AdaLoRA to their full-parameter counterparts alongside DoRA, LoRA+, and VeRA as the state-of-the-art PEFT approaches. These results are visualized in Figure \ref{fig:sota_vis} for vision and Figure \ref{fig:sota_lang} for language. We provide the same results in tabular format in Appendix \ref {app:detailedtable}. LoRA+, AdaLoRA, and DoRA have roughly the same parameter count and obtain similar performance across all datasets and modalities, with LoRA lagging behind. This is expected, as the former three are more recent PEFT methods. As shown in Section \ref{sec:performance}, Partial variants of AdaLoRA and LoRA obtain similar results to their full-parameter versions, meaning they are competitive with SoTA methods at much lower parameter counts. Compared to VeRA, a method aimed at trainable parameter reduction, our approach obtains competitive results while significantly outperforming VeRA on GTSRB and FGVC-Aircraft. This success stems from balancing parameter count and performance by leveraging the pretrained model for fine-tuning capacity. Moreover, VeRA can only modify the number of parameters by changing the residual rank or sharing low-rank matrices across layers. While both can be done using our approach, our method also allows for sparsification of the low-rank matrices themselves. Therefore, using the accuracy threshold discussed in Section \ref{sec:methodology}, another degree of freedom is added to allow for more flexibility in sparsification of PEFT residuals. 

We further present the results of varying this accuracy threshold in Appendix \ref{app:accthreshold}. The performance on Flowers is notable as methods with lower parameter counts achieve better results. We attribute this to overparameterization in other methods causing overfitting, given Flowers' smaller training set. As mentioned in Section \ref{sec:lth}, flow preservation may be a concern during sparsification due to how the non-zero weights are set up for each layer in an isolated manner, independent of the other layers. In Appendix \ref{app:flowpreservation}, we show this is not an issue for our work, due to the additive nature of LoRA residuals.

\subsection{Sweeping through sparsity ratios}
\begin{wrapfigure}[21]{r}{0.4\textwidth}
    \centering
    \includegraphics[width=0.95\linewidth]{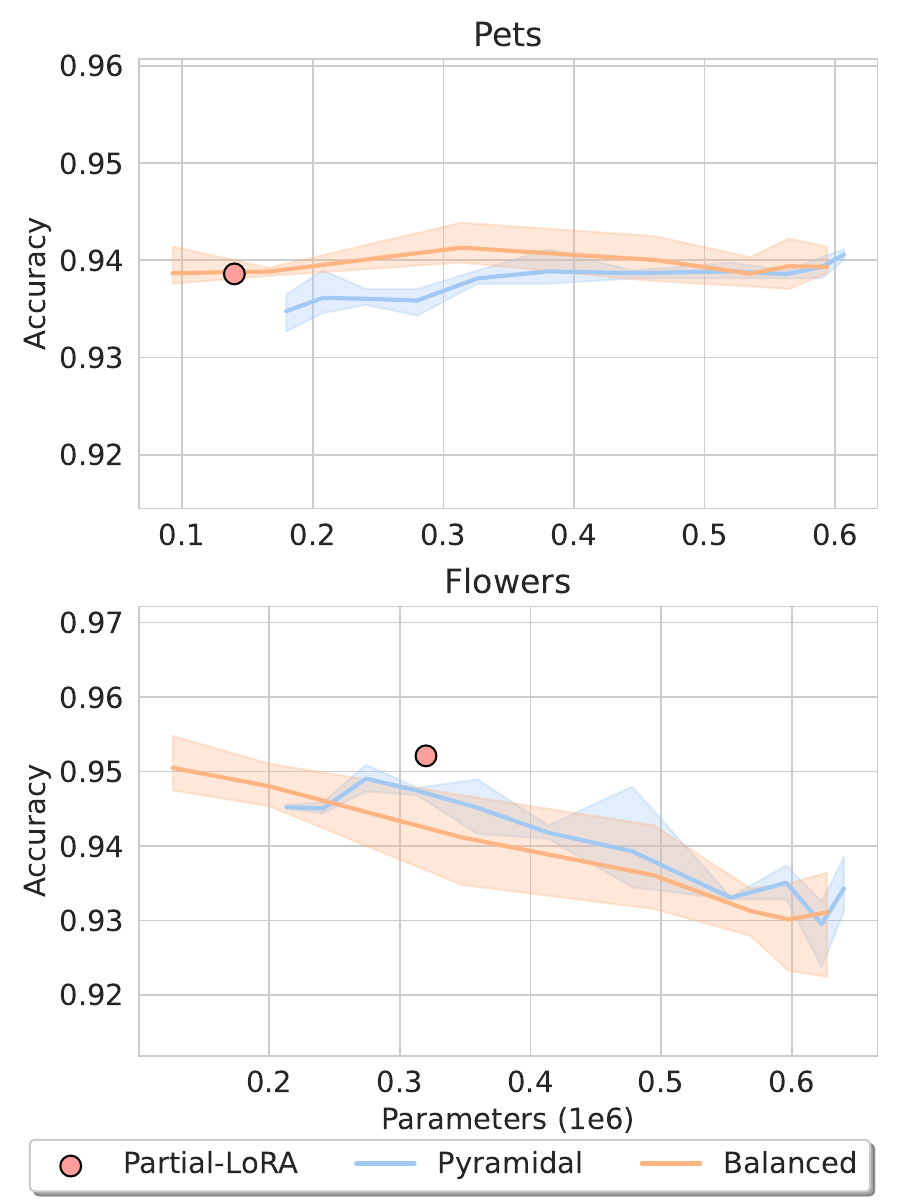}
    \caption{Sweeping sparsity ratio shows sparsification does not lower performance. Partial-LoRA adjusts parameter count to meet dataset demands.}
    \label{fig:sweep}
\end{wrapfigure}

Here, we sweep the sparsity ratios using Balanced and Pyramidal pruning methods to observe performance across different number of parameters. For the Balanced method, we sweep the sparsity ratio over 0.1 to 0.9 and prune every layer with the same sparsity. For Pyramidal pruning, the sparsity is increased exponentially with depth to allow for more parameters at the start and fewer at the later layers. 
As shown in Figure \ref{fig:sweep}, the Pets dataset allows for a consistent reduction in the number of parameters without a significant loss in performance. 
This is while Flowers shows improvements with sparsity which confirms that overparameterization was leading to overfitting as discussed in Section \ref{sec:sota_com}. 

\subsection{Larger Models}
In this section, we study the effectiveness of our approach on LLAMA2-7b, LLAMA3.1-8b and Qwen2.5-14b in a single-task setting. Table \ref{tab:llm} shows that our approach obtains similar accuracy compared to the full fine-tuning and unmasked LoRAs, showing consistency compared to smaller models. While \citep{xu2024randommaskingfindswinning} resorts to sweeping the sparsity ratios to find the 0.001\% sparsity, through our sparsity ratio derivation method (Section \ref{subsec:algorithm}), we obtain a sparsity of around 0.001\% without the need for searching over this parameter or training the mask itself. Moreover, our method, with the dynamic sparsity adapted for each layer, achieves a comparable performance to \citep{xu2024randommaskingfindswinning}. Similar results are obtained in a multi-task setting. As shown in Appendix \ref{sec:multitask}, we sparsify 4 LoRAs to obtain Partial variants trained simultaneously on BoolQ, COPA, RTE, and WiC. Evaluation shows that in a multi-task multi-LoRA setting, sparsified LoRAs obtain the same accuracy as multi-LoRA with significantly smaller parameter count.

\begin{table}[t]
\centering
\caption{Evaluation results across varying model sizes. For Partial-LoRA, parameter counts are listed in parentheses below the accuracy value.}
\label{tab:llm}
\resizebox{0.75\linewidth}{!}{%
\begin{tabular}{lccccc}
\toprule
\multirow{2.5}{*}{\textbf{Model \& Method}} & 
\multirow{2.5}{*}{\textbf{Train. Param.}} & 
\multicolumn{3}{c}{\textbf{Classification}} & 
\multicolumn{1}{c}{\textbf{Multiple Choice}} \\ 
\cmidrule(lr){3-5} \cmidrule(lr){6-6}
 & & \textbf{SST-2} & \textbf{WSC} & \textbf{BoolQ} & \textbf{COPA} \\ 
\midrule

\multicolumn{6}{c}{\textbf{LLAMA2-7B}} \\
\hspace{3mm} FT                    & 6.7B      & 94.7 & 63.5 & 62.2 & 87.0 \\
\hspace{3mm} LoRA                  & 4.2M      & 95.4 & 62.5 & 61.8 & 85.0 \\
\hspace{3mm} Masking (0.001\%)     & 68K       & 95.5 & 64.4 & 58.7 & 88.0 \\
\hspace{3mm} Partial-LoRA (Ours)   & 50K-98K       & \makecell{94.9\\(64K)} & \makecell{66.4\\(50k)} & \makecell{62.0\\(98k)} & \makecell{90.0\\(80K)} \\
\addlinespace

\multicolumn{6}{c}{\textbf{LLAMA3.1-8B}} \\
\hspace{3mm} FT                    & 8B        & 96.0 & 64.9 & 89.3 & 90.0 \\
\hspace{3mm} LoRA                  & 3.4M      & 96.2 & 66.4 & 87.2 & 89.0 \\
\hspace{3mm} Masking (0.001\%)     & 80K       & 95.3 & 64.5 & 85.2 & 76.0 \\
\hspace{3mm} Partial-LoRA (Ours)   & 80K-200K       & \makecell{96.2\\(80k)} & \makecell{66.4\\(200k)} & \makecell{86.7\\(110k)} & \makecell{88.0\\(92k)} \\
\addlinespace

\multicolumn{6}{c}{\textbf{Qwen2.5-14B}} \\
\hspace{3mm} FT                    & 14.7B     & 96.3 & 80.0 & 90.7 & 100.0 \\
\hspace{3mm} LoRA                  & 6.3M      & 95.8 & 76.0 & 90.2 & 100.0 \\
\hspace{3mm} Masking (0.001\%)     & 149K      & 96.1 & 70.2 & 88.6 & 98.0 \\
\hspace{3mm} Partial-LoRA (Ours)   & 92K-400K       & \makecell{96.2\\(92K)} & \makecell{74.1\\(400k)} & \makecell{89.6\\(175k)} & \makecell{100.0\\(122K)} \\
\bottomrule
\end{tabular}%
}
\vspace{-3mm}
\end{table}

\subsection{Sparsity Ratio Derivation Ablation}
\textbf{Partial-LoRA and Targeted-LoRA:} The results in Section \ref{sec:performance} are obtained with Partial-LoRA through random sampling of masks. Alternatively, the specific masks derived by the importance measuring criteria $I^l$ may also be used directly, which we call Targeted-LoRA. This way, regardless of the dataset, the number of parameters for Targeted-LoRAs will be the same as Partial-LoRA. We report the results of Targeted-LoRA in Table \ref{tab:inversion}. On average performance, Targeted-LoRA scores similar to LoRA while slightly falling behind Partial-LoRA. The case-by-case accuracy of both approaches is also close for most datasets. However, for FGVC-Aircraft, Partial-LoRA outperforms Targeted-LoRA by 1\%, with Targeted-LoRA falling behind LoRA by a similar margin.

This shows that targeting specific subnetworks is unnecessary and the layer-specific capacity for the downstream task is the only important factor. We provide a deeper analysis of the behavior of Partial-LoRA and Targeted-LoRA in Appendix \ref{sec:magnitude}. We use the norm of the residuals to show the similarity of Targeted and Partial methods across layers. We show that the changes in magnitude for both these methods across layers, while similar, vary compared to LoRAs signaling potential differences in the fine-tuning dynamic. Additionally, to explore a middle ground between random and deterministic sampling of masks, we treat the importance values as a distribution and sample subnetworks from this distribution. The result of training these subnetworks is provided in Appendix \ref{sec:stochastic} where we show slight improvements in accuracy compared to Partial and Targeted-LoRAs.

\begin{table}[t]
    \centering
    \caption{Results using different sparsity ratio derivation methods. Random masking improves accuracy compared to the deterministic (Targeted) variant, showing how targeted masking is unnecessary.}
    \vspace{0.5mm}
    \label{tab:inversion}
    \resizebox{0.65\textwidth}{!}{
        \begin{tabular}{lc*{5}{>{\centering\arraybackslash}p{1.0cm}}@{}}
        \toprule
        Dataset & LoRA & Targeted-LoRA & \multicolumn{3}{c}{Partial-LoRA} & \multicolumn{1}{c}{Inverted} \\
        & & (SVD) & (SVD) & (SNIP) & (IMP) & \\ \midrule
        Pets          & 94.06 & 93.44 & 93.86 & 93.95 & 93.82 & 94.00 \\
        Flowers       & 93.18 & 95.01 & 95.21 & 95.24 & 94.48 & 93.71 \\
        CIFAR-10      & 94.50 & 94.45 & 94.35 & 94.22 & 93.89 & 94.59 \\
        CIFAR-100     & 80.90 & 80.85 & 81.10 & 80.97 & 80.51 & 80.58 \\
        FER2013       & 57.96 & 58.51 & 58.61 & 58.86 & 57.94 & 58.43 \\
        GTSRB         & 95.04 & 94.64 & 94.72 & 94.53 & 94.90 & 94.40 \\
        FGVC-Aircraft & 54.34 & 53.44 & 54.58 & 55.09 & 54.21 & 51.86 \\ \midrule
        \textbf{Average} & 81.42 & 81.48 & 81.78 & 81.84 & 81.39 & 81.08 \\ \bottomrule
        \end{tabular}}
        \vspace{-5mm}
\end{table}
\textbf{Importance Measuring Criterion:} Aside from the proposed method in Section \ref{sec:importance}, we can also use aforementioned gradient-based importance measuring criterion from the pruning literature to infer the sparsity ratio. The columns named SNIP and IMP under Partial-LoRA in Table \ref{tab:inversion} represent the results of these methods. SVD and SNIP-based methods obtain a similar performance while IMP falls slightly behind due to a lower performance across multiple datasets. To get a deeper insight into the similarity between these methods, we visualize the overlap of the subnetworks extracted by these methods in a pairwise setting. This visualization is provided in Appendix \ref{sec:agreement_importance} where there is a large overlap between masks from the three methods in the first 6 layers of the model. The relatively smaller overlap in the later layers is mainly due to the small size of the subnetwork itself. We also provide a visualization on the relative size of the subnetworks across layers for all three importance measures in Appendix \ref{sec:elementcount}. The initial layers have the smallest sparsity ratio for most datasets while this sparsity grows as we move towards the final layers. For Pets, the sparsity is similar across all layers due to the high performance of the pretrained model before fine-tuning.
\begin{wrapfigure}[21]{RH}[0pt]{0.4\textwidth}
    \centering
    \vspace{-3.5mm}
    \includegraphics[width=0.9\linewidth]{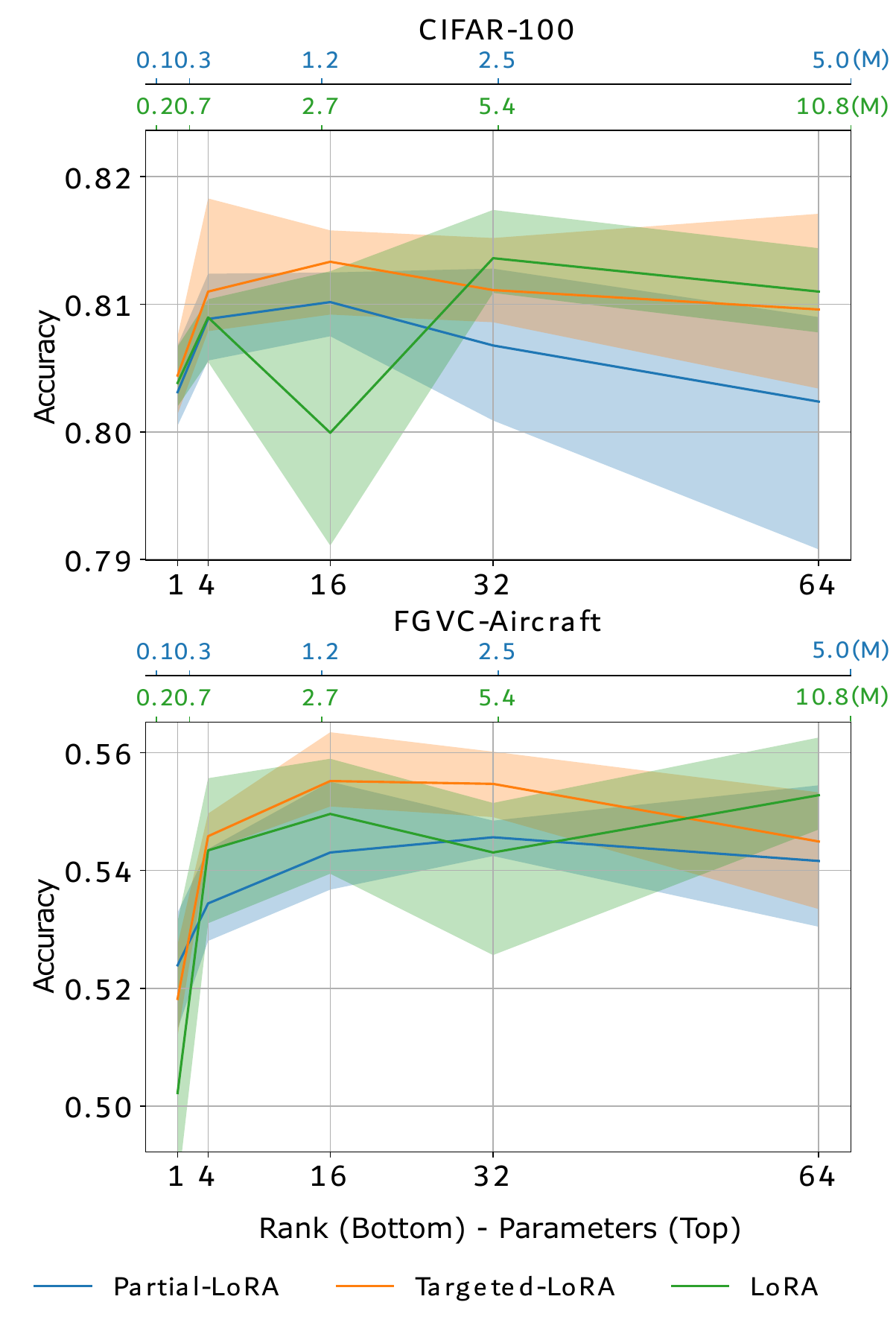}
    \vspace{-2.5mm}
    \caption{Rank ablation with parameter count shows sparsification results stay consistent across varying ranks.}
    \label{fig:rankablation}
 \end{wrapfigure}

\textbf{Inverted Masks:} Our method generates masks that lead to sparse residuals for the adapter. As an ablation, we run the same experiments with these masks inverted. This way, due to the sparsity of Targeted-LoRAs, the inverted masks will have more trainable parameters. A performance improvement here would indicate a larger mask size for Targeted-LoRA is necessary or that Targeted-LoRA misses important elements. The results are reported in Table \ref{tab:inversion}. For Pets and CIFAR-10, the performance of inverted masks is better than Targeted-LoRA, while for the rest, this method results in competitive or degraded performance. Overall, this shows our method identifies and samples the appropriate masks at each layer.

\subsection{Ablation On The LoRA Rank}
\label{sec:rankablation}
Here, we examine the effects of rank on Partial-LoRA. We sweep ranks over 1 to 64, visualizing performance alongside the maximum and minimum accuracy across multiple training sessions in Figure \ref{fig:rankablation} for CIFAR-100 and FGVC-Aircraft, with CIFAR-10 and GTSRB in Appendix \ref{app:rankablation}. Although the top-performing method varies by dataset, the relative behavior from Section \ref{sec:performance} remains consistent. Our method does not degrade performance, showing the results are not dependent on the rank.

\section{Conclusion}
\label{sec:conclusion}
In this study, we demonstrated that random masking in LoRAs can effectively reduce trainable parameters while maintaining performance in fine-tuning large pre-trained models. Drawing from the lottery ticket hypothesis, we showed that sparse "winning ticket" subnetworks within LoRA can match fully parameterized versions, reducing parameters by up to 87\%. Our theoretical framework established that targeted pruning of specific locations is unnecessary and performance is maintained as long as per-layer sparsity remains within our derived bounds. Ablation studies confirmed that flow preservation is unnecessary due to LoRA's residual formulation. Our Partial-LoRA method can be implemented atop existing and future LoRA-based approaches, enabling computational savings while retaining their advantages. This work advances more sustainable deployment of large pre-trained models, with future directions including higher sparsity and analysis of training dynamics.

\bibliographystyle{plainnat}
\bibliography{neurips_2025}


\clearpage
\newpage
\appendix
\newpage
\section{Proof of Theorem 4.1}
\label{app:proof}
\setcounter{theorem}{0}
\renewcommand{\thetheorem}{4.1}
\begin{theorem}
     Define a network $f_{T}$ of depth L parameterized by pretrained weights and biases $W_l$ and $b_l$ with low-rank adapters at each layer parameterized by the residuals $\Delta W^T_l$. Additionally, define a pruned network $f_{LoRA}$ of depth $L+1$ parameterized by $\Delta W_l \cdot U$ and the same $W_l$, $b_l$ as the target model. The pruned model consists of sparsity factors $p_l$ for layer $l$ and $\Delta w^l_{ij} \sim U([-1, 1])$. Both networks have $n_{T, l}$ and $n_{LoRA, l}$ neurons at layer $l$. Then given variables $\epsilon, \delta \in (0, 1)$, with failure probability $1-\delta$, it holds that $\max_{x\in \mathcal{D}}||f_{T}(x; \Delta W^T)-f_{LoRA}(x; \Delta W \cdot U)||\leq \epsilon$ for compact space $D$, if
\end{theorem}
\begin{equation}
    n_{LoRA,l} \geq C\frac{n_{T,l}}{\log(\sfrac{1}{1-p_{l+1}})}\log(\frac{1}{\min(\epsilon_l, \sfrac{\delta}{\rho})}).
\end{equation}
\textbf{Proof}\; We extend the proof by \citep{gadhikar_why_2023} for the existence of strong lottery tickets (SLTs) in Erd\H{o}s-R\'enyi (ER) networks to low-rank adapters. As mentioned in Section \ref{sec:background}, for each layer of a model, after the application of the low-rank adapter the output of the layer previously defined as $f(x) = Wx+b$ becomes $f(x)=(W+\Delta W)x+b$ where $W$ and $b$ are the weight and bias of the pretrained layer and $\Delta W$ represents the learnable residuals. Therefore for both the pruned LoRA and target LoRA we have the following.
\begin{equation}
    f_{T}(x)=(W+\Delta W^T)x+b.
\end{equation}
Whereas for the pruned layer, we have,
\begin{equation}
    f_{LoRA}(x) = (W+\Delta W \cdot U)x+b.
\end{equation}
Where $U$ represents the mask randomly sampled from a Bernoulli distribution. Given that $W$ and $b$ are the same frozen parameters from the pretrained model, and hence will remain unchanged throughout the training and across both LoRA and PaLoRA, we have the following.
\begin{equation}
    f_{T}(x)-f_{LoRA}(x) = \Delta W^T x-(\Delta W \odot U) x.
\end{equation}
The remaining elements of $f_{T}$ and $f_{LoRA}$ can be rephrased as new layers without a bias parameter with the weights of the residuals and $Wx+b$ as the new bias parameter. Therefore according to \citep{gadhikar_why_2023}, 
\begin{equation}
    \mathbb{P}[\max_{x\in \boldsymbol{D}}||g_{T}(x; \Delta W^T)-g_{LoRA}(x;\Delta W\cdot U)||\leq\epsilon]\geq 1-\delta.
\end{equation}
Where $\epsilon, \delta \in (0, 1)$, $g_{T}$ and $g_{LoRA}$ represent the new layers, $\rho=\frac{C N_T^{1+\gamma}}{\log \left(1 /\left(1-\min _l p_l\right)\right)^{1+\gamma}} \log \left(1 / \min \left\{\min _l \epsilon_l, \delta\right\}\right)$ for any $\gamma\geq 0$ with $C$ as a distribution dependent constant. $N_T$ represents the number of non-zero parameters of the model. $\epsilon_l$ is defined as follows \citep{burkholz2022most}.
\begin{equation}
\epsilon_l=\frac{\epsilon}{n_{LoRA, L} L}\left[\left(1+B_{l-1}\right)\left(1+\frac{\epsilon}{L}\right) \prod_{k=l+1}^{L-1}\left(\left\|W^T_{(k)}\right\|_{\infty}+\frac{\epsilon}{L}\right)\right]^{-1}, B_l:=\sup _{x \in \mathcal{D}}\left\|\boldsymbol{x}_{LoRA}^{(l)}\right\|_1 .
\end{equation}
Where $x^l_{LoRA}$ represents the features of the pruned LoRA at layer $l$.
\newpage

\section{Mask Decomposition and Sparsity Ratio}
\label{app:mask_decomp}

In this section, we provide a detailed analysis of the mask decomposition approach used in our implementation and demonstrate its theoretical equivalence to the formulation presented in Theorem \ref{th:theorem}. We show that our efficient implementation preserves the theoretical guarantees while reducing computational overhead.

\subsection{Mask Decomposition and Implementation}
In Theorem \ref{th:theorem}, we establish bounds on the outputs of pruned and unpruned LoRAs using a mask in the form $\Delta\mathbf{W} \cdot \mathbf{U}$, where $\mathbf{U}$ is sampled from a Bernoulli distribution $\mathcal{B}(p)$. A direct implementation of this formulation would be computationally inefficient, as it would require computing all parameters of the low-rank residual only to zero out the majority through masking. Instead, we propose an efficient implementation that achieves the same theoretical guarantees by decomposing the mask operation.

Our implementation modifies the formulation of low-rank residuals by zeroing out rows of matrix $\mathbf{B}$ and columns of matrix $\mathbf{A}$ in the LoRA formulation, resulting in $\mathbf{W}_{masked} = \mathbf{U}_r\mathbf{B}(\mathbf{U}_c\mathbf{A}^\top)^\top$. While matrix multiplication is non-commutative, this formulation yields equivalent sparsity ratios to $\Delta\mathbf{W} \cdot \mathbf{U}$. This is because the key constraint on $\mathbf{U}$ in Theorem \ref{th:theorem} is the sparsity ratio $p_l$ at each layer. This sparsity can be induced through the decomposed masks $\mathbf{U}_r$ and $\mathbf{U}_c$. We can express the relationship between the masked and decomposed formulations for each element in row $i$ and column $j$ as follows:
\begin{align}
M_{ij}^{decomposed} &= \mathbf{U}_r\mathbf{B}(\mathbf{U}_c\mathbf{A}^\top)^\top = \sum_{k=1}^d (\mathbf{U}_r)_{ii}B_{ik}(\mathbf{U}_c)_{jj}A_{kj}, \label{eq:decomposedformulation} \\
M_{ij}^{masked} &= (\mathbf{B}\mathbf{A})_{ij} \mathbf{U}_{ij} = \sum_{k=1}^d B_{ik}A_{kj}\mathbf{U}_{ij}.
\end{align}
In expectation, (\ref{eq:decomposedformulation}) has a sparsity ratio ($p$) bounded through the Bernoulli distribution in Theorem \ref{th:theorem}. This value can be easily preserved using the decomposed formulation due to the following:
\begin{equation}
\mathbb{E}[\mathbf{U}_{ij}] = \mathbb{E}[(\mathbf{U}_r)_{ii}(\mathbf{U}_c)_{jj}] = p \quad \forall i,j.
\end{equation}
Where the sparsity ratio of $\mathbf{U}_r$ and $\mathbf{U}_c$ can be modified arbitrarily.

\subsection{Practical Implications}
Our implementation initially used the non-decomposed formulation, and upon switching to the decomposed implementation, we observed equivalent performance in our vision experiments. The decomposed implementation provides significant computational advantages on modern GPU architectures while preserving all theoretical guarantees.

\subsection{Adjusting the Accuracy Threshold}
\label{app:accthreshold}
Here, we present an analysis of the performance of Partial-AdaLoRA under varying accuracy thresholds, specifically adjusting the 90\% margin in Section \ref{sec:importance} to 70\% and 80\%. The evaluation was conducted on the Flowers and GTSRB datasets. The Flowers dataset was selected due to its higher performance compared to unpruned LoRA, while GTSRB was chosen as it demonstrated superior performance compared to VeRA. The results are presented in Table \ref{tab:accthreshold}.

\begin{table}[h]
\centering
\caption{Partial-AdaLoRA accuracy across different accuracy thresholds}
\resizebox{0.55\textwidth}{!}{
\begin{tabular}{lccc}
\toprule
Dataset & 70\% & 80\% & 90\% \\
\midrule
GTSRB & 93.11 (0.14M) & 93.61 (0.16M) & 94.72 (0.26M) \\
Flowers & 94.32 (0.11M) & 94.05 (0.17M) & 95.21 (0.32M) \\
\bottomrule
\end{tabular}}

\label{tab:accthreshold}
\end{table}

On GTSRB, we maintain the performance gap relative to VeRA with the reduction in the number of parameters while achieving similar parameter count to VeRA. On Flowers, the accuracy drops a percentage point when lowering the 90\% threshold. This shows how suitable the 90\% margin is in striking a balance between parameterization and performance.

\section{From Low Data Regime to Training on the Whole Dataset}
\label{sec:wholedataset}
As shown in section \ref{sec:performance}, the number of trainable parameters is significantly reduced for for Partial-LoRAs. Here, we assess if the performance similarity between LoRAs and Partial-LoRAs is due to the large number of parameters relative to the small training dataset size. Table \ref{tab:fulldatasettraining} shows the performance of each method compared to LoRAs when trained on the entire training set $\mathcal{D}$ instead of a few-shot subset $\mathcal{D}_t$. The performance similarity between the three methods in Table \ref{tab:fulldatasettraining} is consistent with the few-shot results in section \ref{sec:performance}. Therefore, pruned LoRAs perform similarly to LoRAs, regardless of dataset size.
\begin{table}[h]
    \centering
    \caption{Performance of training on the whole dataset}
    \label{tab:model_performance}
    \begin{center}
    \resizebox{0.55\textwidth}{!}{
    \centering
    \begin{tabular}{@{}lccc@{}}
    \toprule
    Dataset         & LoRA  & Targeted-LoRA  & Partial-LoRA     \\ \midrule
    CIFAR-10        & 96.68 & 97.68   & 97.79  \\
    CIFAR-100       & 87.16 & 87.47   & 87.59  \\
    GRSRB           & 99.12 & 99.02   & 99.08  \\
    FGVC-Aircraft   & 65.61 & 64.93   & 65.65  \\
    Average         & 87.14 & 87.28   & 87.53  \\ 
    \bottomrule
    \end{tabular}}
    \end{center}
\label{tab:fulldatasettraining}
\end{table}

\section{Multi-Task Learning within a Multi-LoRA Setup}
\label{sec:multitask}
Here we assess the effects of our Partial-LoRA approach in a setting where multiple LoRAs are used. We use datasets BoolQ, COPA, RTE, and WiC for this experiment. We use a few-shot subset of each dataset's training set with 10\% of the available samples to finetune the model. Four LoRAs are simultaneously trained on the concatenated datasets and then evaluated on each dataset to form Table \ref{tab:multilora}. These results are named Multi-LoRA. By using the same few-shot training set, we extract sparsity ratios for each dataset based on the pretrained base model LLAMA2-7b. Then, we sparsify one of the LoRAs for each dataset, resulting in a Partial-Multi-LoRA model. Each LoRA has a rank of 24 and we set the scaling parameter $\alpha$ for each LoRA to 48. We train both models with the same hyperparameters except for the learning rate which we set to $1e-3$ for Partial-Multi-LoRA and $2e-4$ for Multi-LoRA which we found through a sweep in the range of $1e-2$ and $1e-5$. We also compare the results of both Multi-LoRA and Partial-Multi-LoRA with LoRA itself in a setting where the adapter for LoRA has a rank of 96. 
\begin{table}[h]
\centering
\caption{In a multi-LoRA setting with 4 LoRAs where each LoRA is sparsified using our Partial-LoRA method, we obtain similar results to the dense counterpart. This is while a few-shot subset of the datasets is concatenated and used for training in a multi-task setting. Parameter counts are shown in parentheses.}
\vspace{2mm}
\label{tab:multilora}
\resizebox{0.78\linewidth}{!}{
\begin{tabular}{lccccc}
\toprule
\textbf{Method} & \textbf{BoolQ} & \textbf{COPA} & \textbf{RTE} & \textbf{WiC} & \textbf{Avg.} \\ 
\midrule
LoRA & 87.0 (100M) & 93.0 (100M) & 87.3 (100M) & 71.1 (100M)  & 84.6 (100M)\\
Multi-LoRA   & 87.8 (100M) & 91.0 (100M) & 87.3 (100M) & 72.1 (100M) & 84.55 (100M) \\
Partial-Multi-LoRA  & 87.5 (20M)& 90.0 (20M) & 86.0 (20M) & 74.2 (20M) & 84.42 (20M) \\
\bottomrule
\end{tabular}}
\end{table}
By sparsifying LoRAs with a slightly loosened accuracy margin, we obtain a Partial-Multi-LoRA setup with about 20\% of the trainable parameters of the Multi-LoRA and LoRA approaches. Despite the lower parameter count, we obtain similar results to both LoRA and Multi-LoRA on average. Interestingly, we obtain better results on a more difficult dataset WiC, while slightly falling behind Multi-LoRA on COPA. We believe this is due to COPA and WiC's dataset size and difficulty. While small in number of training samples, COPA can take advantage of the model's training on the other datasets due to its lower difficulty. On the other hand, WiC, being a difficult task, requires specific subspaces/LoRAs to be trained for obtaining a good performance on this dataset. Our masking approach could potentially reduce the overlap between masks, leading to isolated parameters that can adjust different parts of the weight matrix, resulting in better WiC performance. This shows a potential for future work where different parts of each weight matrix are assigned randomly to different datasets to reduce interference with shared sections for collaboration between datasets.

\section{Magnitude of Residuals}
\label{sec:magnitude}
\begin{figure}[h]
\centering
\includegraphics[width=0.85\textwidth]{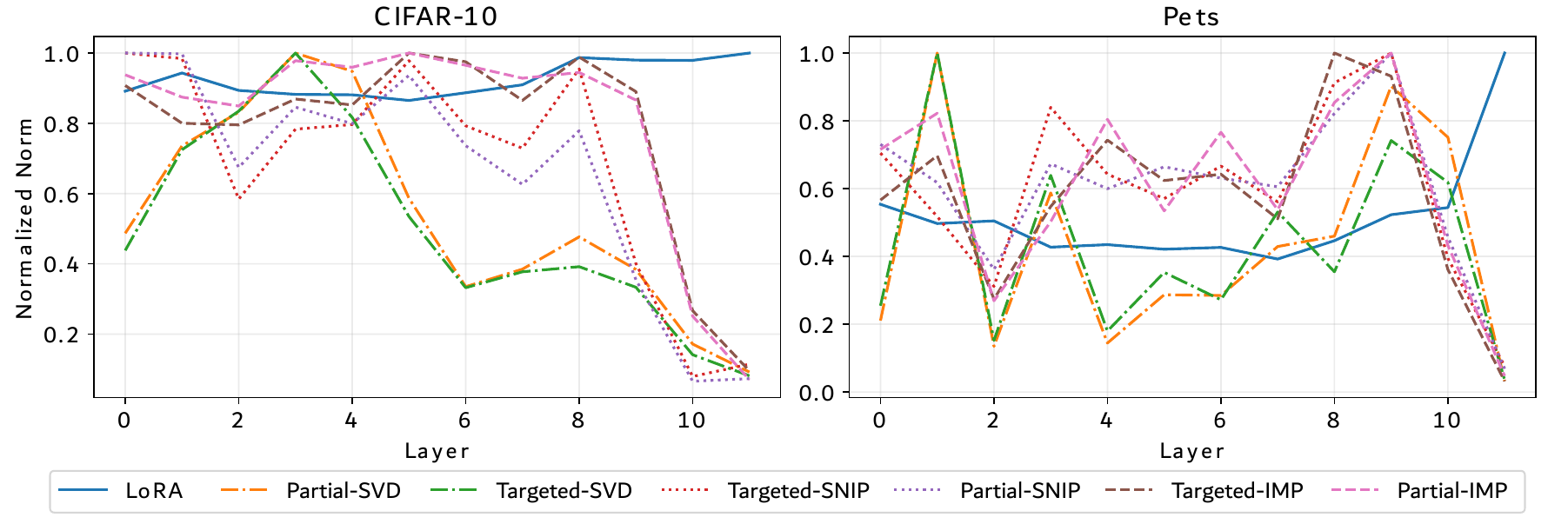}
\caption{Normalized magnitude of residualsfor CIFAR-10 and Pets for the first fully-connected layer of every transformer block.}
\label{fig:main_performance_normed}
\end{figure}

\begin{figure}[h]
\centering
\includegraphics[width=0.85\textwidth]{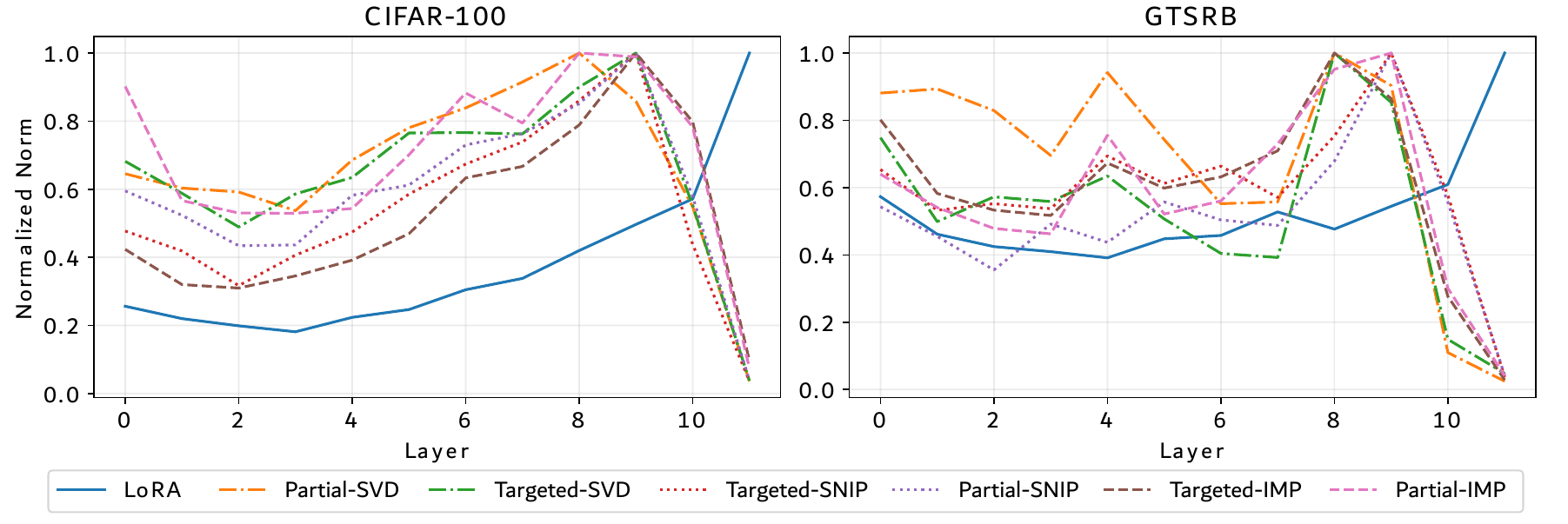}
\caption{Magnitude of the residuals for datasets CIFAR-100 and GTSRB for the first fully-connected layer of every transformer block. The magnitude goes down as layer depth increases.}
\label{fig:main_performance2}
\end{figure}

\begin{figure}[h]
\centering
\includegraphics[width=0.85\textwidth]{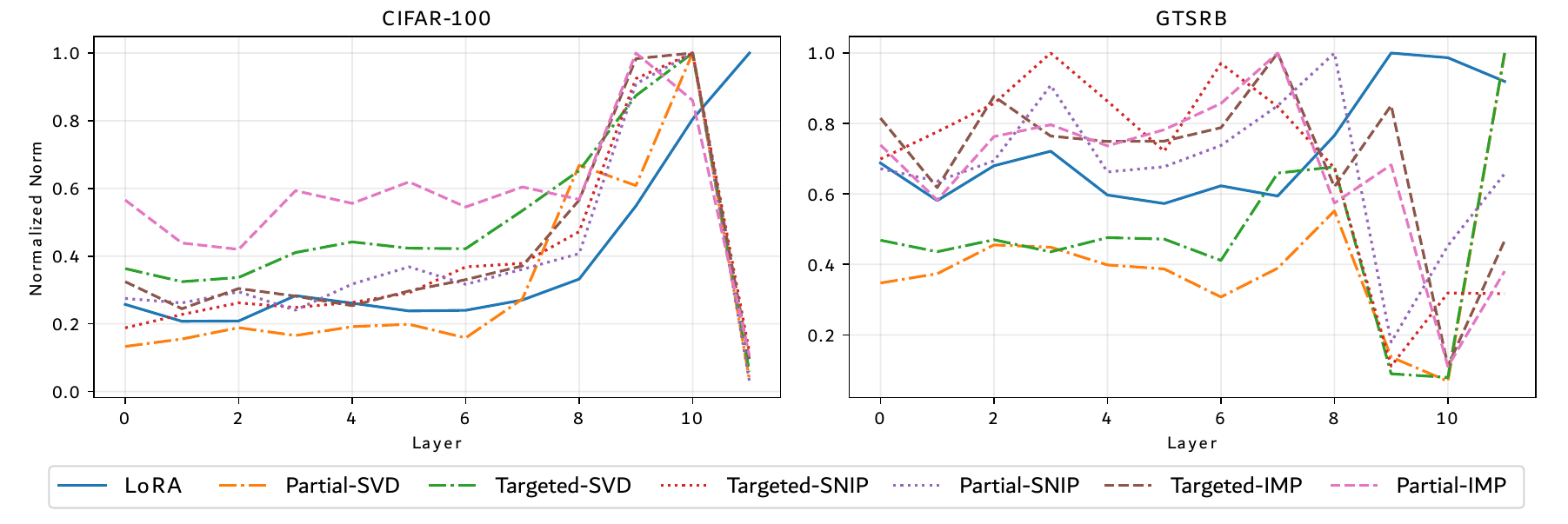}
\caption{Magnitude of the residuals for datasets CIFAR-100 and GTSRB for the second fully-connected layer of every transformer block. The general pattern follows that of the first layer of the transformer block.}
\label{fig:magnitude_cproj}
\end{figure}

\begin{figure}[h]
\centering
\includegraphics[width=0.85\textwidth]{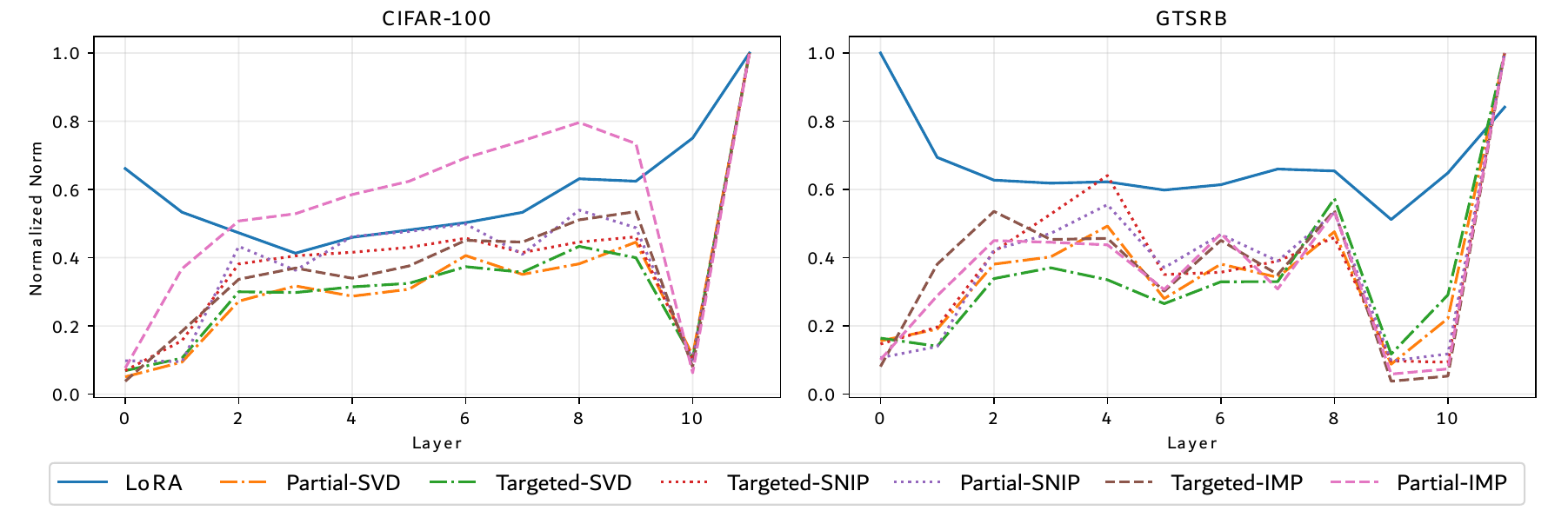}
\caption{Magnitude of the residuals for datasets CIFAR-100 and GTSRB for the attention layer (inward projection) of every transformer block. The pattern is opposite of that of the fully-connected layers. This could be due to how the attention layer is itself made of 3 layers, namely, the query, key and value projections.}
\label{fig:magnitude_attn}
\end{figure}

While LoRAs modify all weights, Partial-LoRAs modify only a small subset. We study the impact of this reduction in the number of trainable parameters on the magnitude of the residuals obtained through fine-tuning. Figure \ref{fig:main_performance_normed} shows the norm of the residual matrix across the 12 layers of the transformer model, focusing on the first fully connected layer of each block, with additional visualizations for CIFAR-100 and GTSRB datasets visualized in Figure \ref{fig:main_performance2}. Moreover, Figure \ref{fig:magnitude_cproj} and Figure \ref{fig:magnitude_attn} visualize the same norms for the second fully connected layer and attention projection layers, respectively. Alongside Partial-LoRA and Targeted-LoRA derived using our proposed SVD-based method, we also provide the visualization for SNIP and IMP based Targeted-LoRAs as well.

The norm of the residual matrix from Targeted-LoRA behaves similarly to the Partial-LoRA method across all importance measures, indicating the importance of the number of modified parameters over the specific elements. Additionally, we note differences between LoRAs and masked LoRAs. Neither masking approach achieves magnitudes similar to LoRAs, changing the model's overall behavior. For example, LoRA shows an uptick in norm for the final layer of the Pets dataset, while masked models show the opposite. In CIFAR-10, LoRA maintains consistent norm values across layers, whereas masked LoRAs exhibit drastic changes. This pattern remains consistent across the datasets with the norm value going down significantly for the final layer. Additionally, the norm for Partial-LoRA follows the Targeted method closely across each layer for every importance measure. This gives us insights on how the randomly masked LoRAs bearing the same capacity as the Partial LoRAs can modify the pretrained weights in a similar manner.

Here we provide the visualization of the residual norms for two other datasets that are CIFAR-10 and GTSRB. 

\section{Exploring Deterministic and Stochastic Utilization of Importance Measures}
\label{sec:stochastic}
\begin{figure}[h]
\centering
    \includegraphics[width=0.8\textwidth]{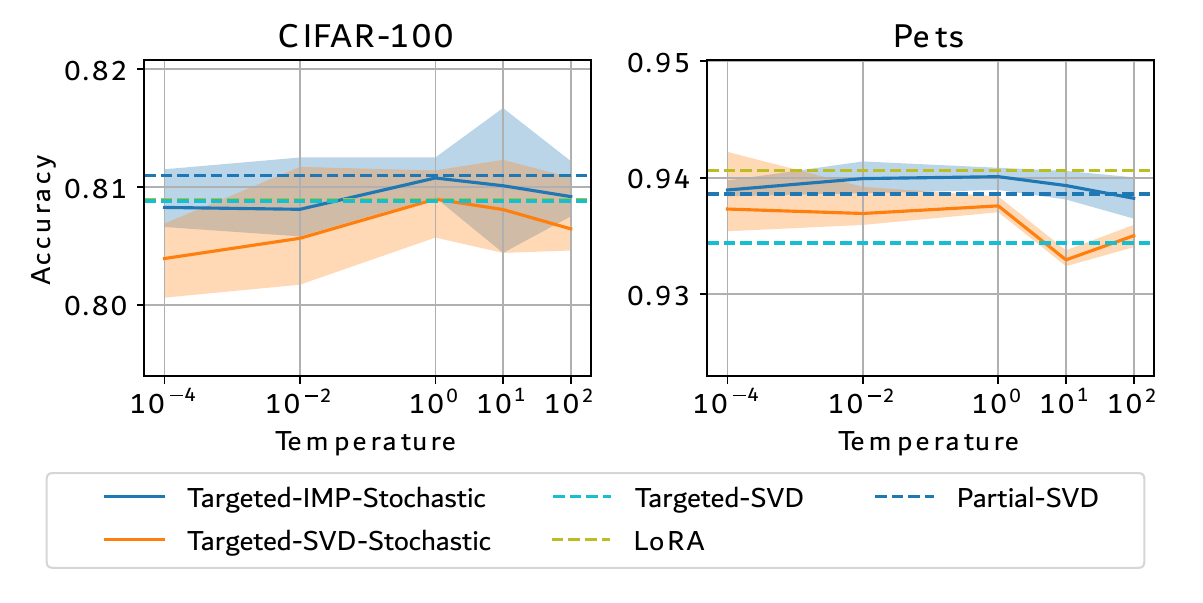}
    \caption{Stochastic approach to subnetwork extraction has minimal effects on performance.}
    \label{fig:stochastic}
\end{figure}

In Section \ref{sec:performance}, Targeted-LoRA methods used deterministic approaches to select elements for modification based on importance scores. Here, we treat these scores probabilistically, sampling indices for modification. We introduce randomness by passing importance values through a softmax function and explore the effects by varying the temperature parameter. Then, we sample masks using the derived distribution. Table \ref{fig:stochastic} shows that the stochastic IMP approach achieves accuracy close to LoRAs for Pets and improves performance on CIFAR-100 across temperatures. However, the SVD-based method's performance degrades with randomness, possibly due to its dependence on the SVD of the pretrained matrix, unlike the IMP method, which extracts scores from gradient magnitudes, allowing for a more meaningful interpretation of this stochasticity.

\section{Agreement in the Sub-Network Extraction.}
\label{sec:agreement_importance}

In this section, we aim to determine the similarity between sub-networks extracted by SVD, SNIP, and IMP. We hypothesize that if there is a meaningful set of elements in each weight matrix responsible for a model's performance, there would be some overlap between the sub-networks identified by these different methods. The results for CIFAR-10, Pets, CIFAR-100 and GTSRB are provided in the following.

\begin{figure}[h]
\centering
\includegraphics[width=0.999\textwidth]{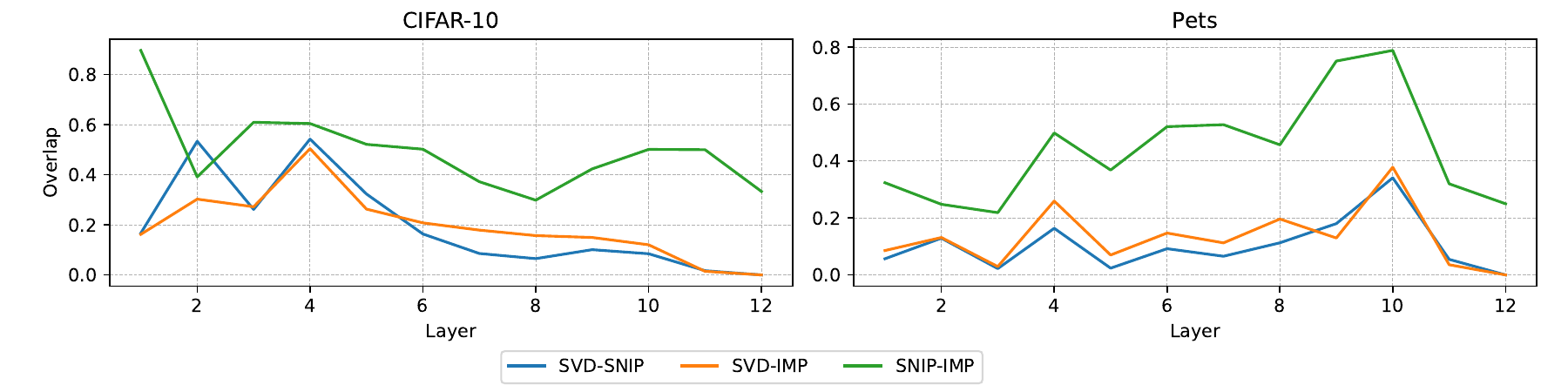}
\caption{Overlap of top indices across different importance measures for CIFAR-10 and Pets. }
\label{fig:methodoverlap2}
\end{figure}

\begin{figure}[h]
\centering
\includegraphics[width=0.67\textwidth]{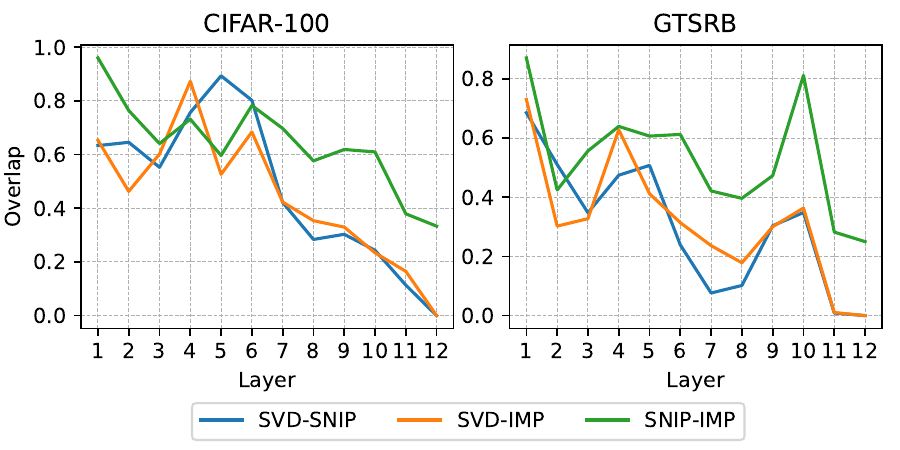}
\caption{Overlap of top indices across different importance measures for CIFAR-100 and GTSRB. }
\label{fig:methodoverlap3}
\end{figure}

\textbf{Overlap of layers on different methods:}
In Figure \ref{fig:methodoverlap2}, the overlap for CIFAR-10 and Pets is larger than 50\% in the first 6 layers of the model when we compare the sub-network extracted by SVD-based and gradient-based methods. This suggests that using importance measures to infer the number of elements to modify is effective, highlighting the similarity between the SVD-based and gradient-based methods. Similar results are shown in Figure \ref{fig:methodoverlap3} for CIFAR-100 and GTSRB. Interestingly, the overlap between the SVD-based method and the gradient-based methods is different from the other datasets. For Pets, the overlap is consistently low across the layers. This could potentially be due to the fine granular information required to be modified by the LoRA. If so, a larger portion of the top singular vectors would be used for the subnetwork extraction phase, leading to a smaller overlap across all layers.

\begin{figure}[h]
\centering
\includegraphics[width=0.999\textwidth]{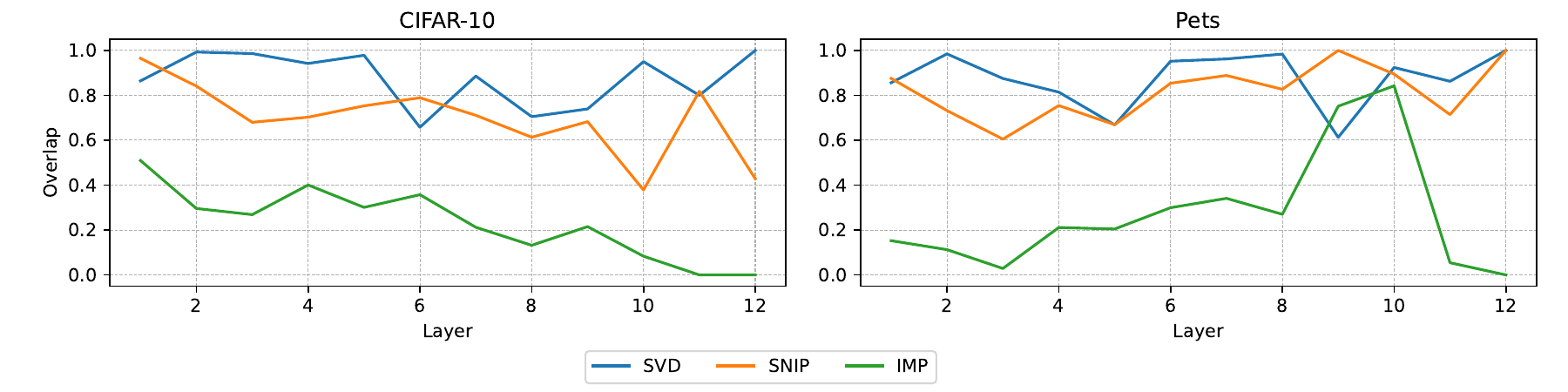}
\caption{Overlap of top indices across different shots for CIFAR-10 and Pets.}
\label{fig:shotoverlap2}
\end{figure}

\begin{figure}[h]
\centering
\includegraphics[width=0.67\textwidth]{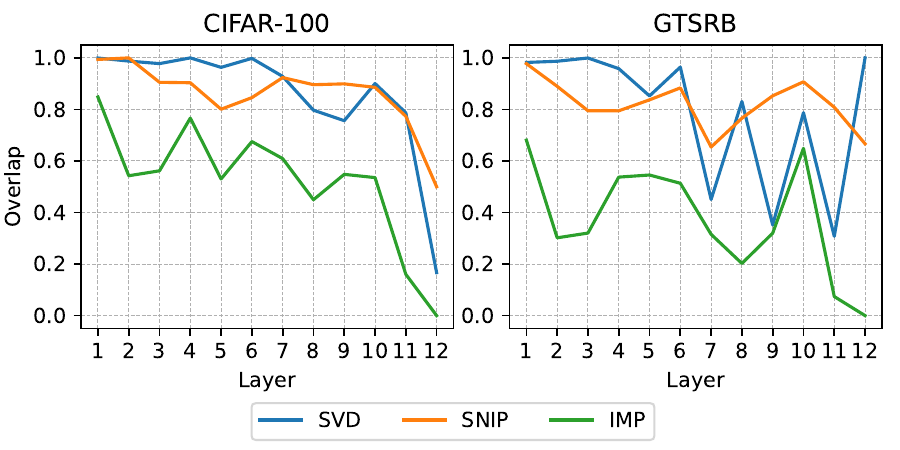}
\caption{Overlap of top indices across different shots for CIFAR-100 and GTSRB.}
\label{fig:shotoverlap4}
\end{figure}

\textbf{Overlap of layers on the same method but different shots:}
We compare each method using different shots from each dataset to complete these observations. Shared concepts should be reflected in overlaps of sub-networks across different shots. As shown in Figure \ref{fig:shotoverlap2}, the SVD-based method shows a large overlap between different seeds, with full overlap for GTSRB across all layers. This is expected since SVD starts with the top singular vectors. Surprisingly, the gradient-based method also shows significant overlap across different shots for both datasets.
Therefore, considering the results in Table \ref{tab:inversion}, while specific indices may not be optimal for training, the sparsity factor can be effectively determined using importance measuring methods, as shown by the comparisons between different importance methods. Similar results are shown in Figure \ref{fig:shotoverlap4} for CIFAR-100 and GTSRB. These results show that the sparsity ratio is not dependent on the specific samples passed through the model and while the specific indices chosen for the mask might change, the sparsity ratio itself is mostly consistent across further validating our proposed approach in Section \ref{sec:importance}.

\section{Rank ablation study for other datasets}
\label{app:rankablation}
\begin{figure}[h]
\centering
\includegraphics[width=0.75\textwidth]{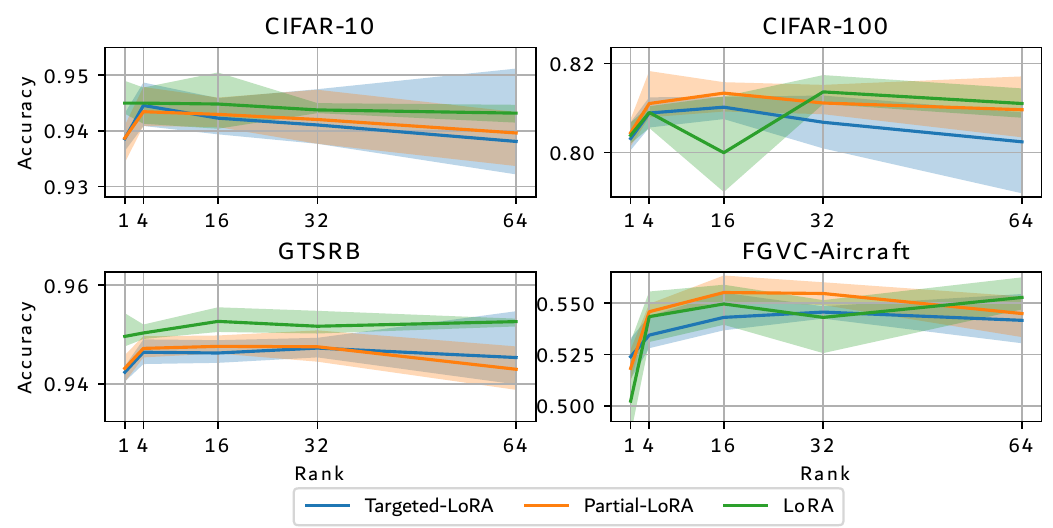}
\caption{Ablation on the rank of the low-rank residual weights. Regardless of rank, random masking allows for reduction in the number of parameters without significant reduction in performance.}
\label{fig:rankablationappendix}
\end{figure}
In this section, we provide the visualizations for the ablation study on the rank of the low-rank residuals for CIFAR-10 and GTSRB. Fig. \ref{fig:rankablationappendix}, visualizes the accuracy values across ranks of 1, 4, 16, 32 and 64. Similar to the results for CIFAR-100 and FGVC-Aircraft, the relative performance between LoRAs and pruned counterparts is consistent showing that performance of pruned LoRAs is not dependent on rank.

\begin{figure}[h]
\centering
\includegraphics[width=0.75\textwidth]{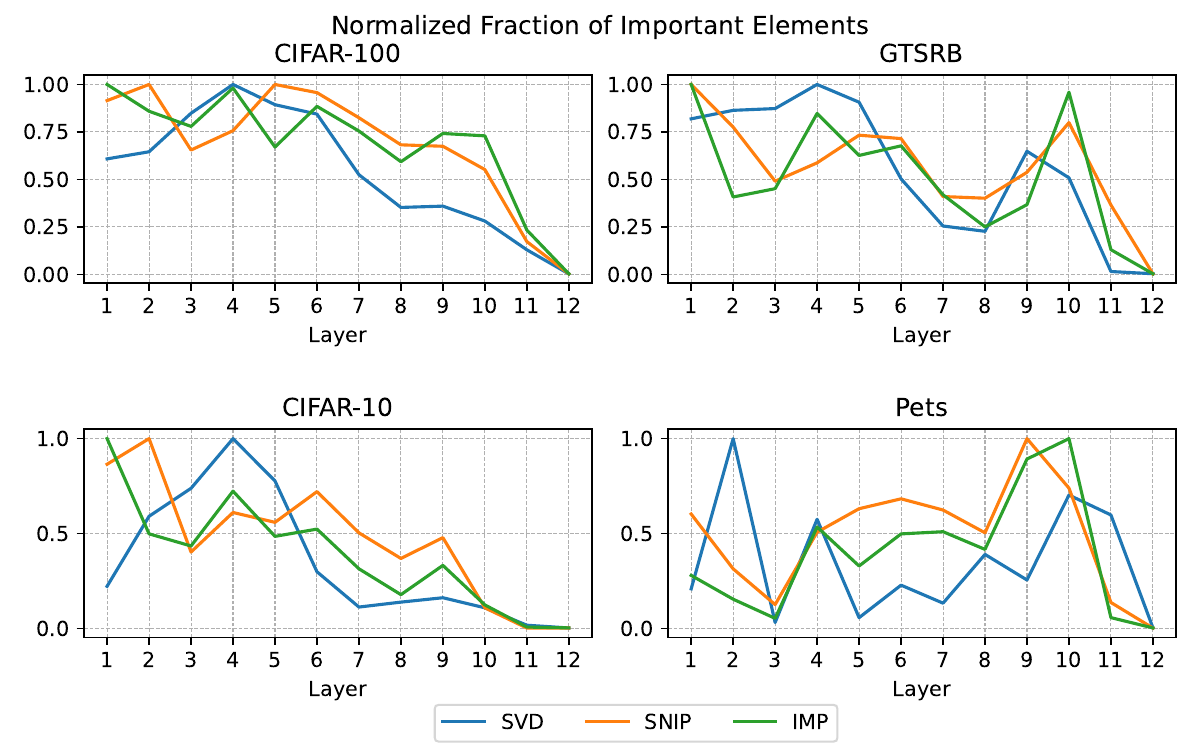}
\caption{Fraction of the number of elements in the extracted subnetwork from different importance measures over the number of parameters in the pretrained model. The number of important elements goes down in deeper layers suggesting narrowing of the high-level knowledge required to infer accurately on a task.}
\label{fig:fractionelements}
\end{figure}

\section{Important elements across different layers}
\label{sec:elementcount}

In this section, we visualize the fraction of elements in the subnetwork extracted by SNIP, SVD and IMP across the layers. These visualizations are provided in Figure \ref{fig:fractionelements}. As mentioned in Appendix \ref{sec:agreement_importance}, the number of elements chosen to be preserved in the extracted subnetwork decreases as depth increases, causing a smaller overlap between the importance measures.

\section{Effects of flow preservation}
\label{app:flowpreservation}

In the pruning literature, it is common to modify the masks resulting from importance measures to preserve the flow of information throughout the layers of the model. This is due to the possibility that if two layers are pruned independently of each other, the activations resulting from one layer might encounter weights that have been masked out, leading to a vector of zeros as the output of the second layer. Here we show that in the fine-tuning setting where LoRAs adjust the pretrained layers, this step is not necessary. We use the IMP importance method and infer the masked elements for each layer independent of other layers. Then, we use the same approach but for each layer, the calculated scores are added to the scores of the next layer to reweigh the importance of each element in favor of important elements from the next layer. The performance of both methods is reported in Table \ref{tab:flow}. The results with flow preservation are named Continuous and the results without this alteration of the importance measure are named Isolated. Both the individual datasets and the overall average are consistently close confirming that flow preservation is not necessary in the case of our work.

\begin{table}[h]
    \centering
    \caption{Comparison of Isolated and Continuous Data Accuracy}
    \label{tab:flow}
    \resizebox{0.5\textwidth}{!}{
    \begin{tabular}{@{}lcc@{}}
    \toprule
    Dataset        & Isolated (\%) & Continuous (\%) \\ \midrule
    Pets           & 93.82         & 93.64           \\
    CIFAR-10        & 93.89         & 94.20           \\
    CIFAR-100       & 80.51         & 80.86           \\
    GTSRB          & 94.90         & 94.75           \\
    FGVC-Aircraft  & 54.21         & 54.67           \\
    Average           & 83.47         & 83.58           \\ \bottomrule
    \end{tabular}}
\end{table}

\section{Detailed Results from Quantitative Evaluation}
\label{app:detailedtable}

Here we provide the same results from Figure \ref{fig:sota_vis} and Figure \ref{fig:sota_lang} in tabular format. Table \ref{tab:sota_vis} and Table \ref{tab:sota_lang} show the performance of each method on each dataset tested in our work.

\begin{table}[ht]
\centering
\caption{All Performance Results on Vision Tasks with Exact Parameter Count}
\resizebox{0.99\textwidth}{!}{
\begin{tabular}{lcccccccc}
\toprule
\textbf{Method} & \textbf{Pets} & \textbf{Flowers} & \textbf{CIFAR10} & \textbf{CIFAR100} & \textbf{FER2013} & \textbf{GTSRB} & \textbf{FGVC-Aircraft} & \textbf{ImageNet} \\
\midrule
LoRA                 & 94.06 (0.67M) & 93.18 (0.67M) & 94.50 (0.67M) & 80.90 (0.67M) & 57.96 (0.67M) & 95.04 (0.67M) & 54.34 (0.67M) & 71.01 (0.67M) \\
Partial-LoRA         & 93.86 (0.14M) & 95.21 (0.32M) & 94.35 (0.17M) & 81.10 (0.30M) & 58.61 (0.21M) & 94.72 (0.26M) & 54.58 (0.31M) & 70.90 (0.23M) \\
VeRA                 & 94.23 (0.10M) & 94.91 (0.10M) & 95.23 (0.10M) & 81.60 (0.10M) & 58.81 (0.10M) & 91.23 (0.10M) & 52.29 (0.10M) & 73.56 (0.10M) \\
LoRA+                & 94.07 (0.67M) & 94.35 (0.67M) & 94.22 (0.67M) & 81.28 (0.67M) & 59.51 (0.67M) & 93.03 (0.67M) & 51.46 (0.67M) & 72.31 (0.67M) \\
DoRA                 & 93.55 (0.77M) & 93.64 (0.77M) & 94.78 (0.77M) & 81.19 (0.77M) & 59.10 (0.77M) & 94.62 (0.77M) & 53.41 (0.75M) & 71.60 (0.75M) \\
AdaLoRA              & 93.91 (0.67M) & 94.36 (0.67M) & 94.72 (0.67M) & 81.54 (0.67M) & 59.32 (0.67M) & 94.13 (0.67M) & 52.74 (0.67M) & 72.68 (0.67M) \\
Partial-AdaLoRA      & 93.38 (0.12M) & 94.62 (0.26M) & 94.11 (0.15M) & 81.02 (0.29M) & 58.77 (0.21M) & 93.79 (0.25M) & 52.22 (0.30M) & 72.79 (0.22M) \\
\bottomrule
\end{tabular}}
\label{tab:sota_vis}
\end{table}

\begin{table}[ht]
    \centering
    \caption{Performance Comparison Across Language Tasks (Parameter Count)}
    \resizebox{0.8\textwidth}{!}{
    \begin{tabular}{lcccccc}
        \toprule
        \textbf{Method} & \textbf{SST-2} & \textbf{QNLI} & \textbf{MRPC} & \textbf{CoLA} & \textbf{QQP} & \textbf{MNLI} \\
        \midrule
        LoRA                     & 94.95 (1.33M) & 93.87 (1.33M) & 89.95 (1.33M) & 69.82 (1.33M) & 91.99/89.38 (1.33M) & 90.65/90.69 (1.33M) \\
        Partial-LoRA             & 95.60 (0.18M) & 93.85 (0.28M) & 89.87 (0.14M) & 70.86 (0.31M) & 91.76/89.07 (0.23M) & 89.50/89.69 (0.37M) \\
        Partial-AdaLoRA          & 96.00 (0.20M) & 94.31 (0.33M) & 90.68 (0.17M) & 71.63 (0.34M) & 91.93/89.36 (0.24M) & 90.10/90.00 (0.37M) \\
        Vera                     & 95.64 (0.10M) & 94.05 (0.10M) & 90.44 (0.10M) & 71.28 (0.10M) & 91.47/88.65 (0.10M) & 89.35/89.43 (0.10M) \\
        LoRA+                    & 95.98 (1.33M) & 93.87 (1.33M) & 90.60 (1.33M) & 70.85 (1.33M) & 92.11/89.58 (1.33M) & 90.49/90.53 (1.33M) \\
        DoRA                     & 96.10 (1.41M) & 94.14 (1.41M) & 90.68 (1.41M) & 72.11 (1.41M) & 92.21/89.61 (1.41M) & 90.31/90.40 (1.41M) \\
        AdaLoRA                  & 96.10 (1.32M) & 94.55 (1.32M) & 90.69 (1.32M) & 71.45 (1.32M) & 92.23/89.74 (1.32M) & 90.76/90.79 (1.32M) \\
        \bottomrule
    \end{tabular}}
    \label{tab:sota_lang}
\end{table}


\end{document}